\documentclass[10pt,twocolumn,letterpaper]{article}

\usepackage[pagenumbers]{cvpr} 

\usepackage{graphicx}
\usepackage{amsmath}
\usepackage{amssymb}
\usepackage{booktabs}
\usepackage{enumitem}
\usepackage{amsfonts}               
\usepackage{microtype}              
\usepackage{nicefrac}               
\usepackage[dvipsnames]{xcolor}     
\usepackage{url}                    
\usepackage{etoolbox}
\usepackage{xparse}
\usepackage{csquotes}
\usepackage{wrapfig}
\usepackage{bm}
\usepackage{soul}
\usepackage{siunitx}
\usepackage{textcomp}
\usepackage[pagebackref,breaklinks,colorlinks]{hyperref}
\newcommand{\figcolor}[1]{{\color{red}#1}}
\usepackage[capitalize]{cleveref}  
\makeatletter\@namedef{ver@everyshi.sty}{}\makeatother  
\usepackage{tikz}
\usepackage[accsupp]{axessibility}  

\newtoggle{nographics}
\newtoggle{lrgraphics}
\newtoggle{nocomments}
\newtoggle{separatesupp}

\togglefalse{nographics}
\togglefalse{lrgraphics}
\togglefalse{nocomments}
\toggletrue{separatesupp}

\crefname{section}{Section}{Sections}
\crefname{table}{Table}{Tables}
\crefname{figure}{Figure}{Figures}

\togglefalse{separatesupp} 

\makeatletter
\renewcommand{\paragraph}{%
    \@startsection{paragraph}{4}%
    {\z@}{0ex \@plus 0.2ex \@minus .2ex}{-1em}%
    {\normalfont\normalsize\bfseries}%
}
\makeatother

\setlist{topsep=0pt,itemsep=-5pt,leftmargin=*}
\begin{document}
    \def\cvprPaperID{8256} 
\def\confName{CVPR}
\def\confYear{2023}

\title{Multi-sensor large-scale dataset for multi-view 3D reconstruction}

\author{\
\hfill{}Oleg Voynov\textsuperscript{1,2}
\hfill{}Gleb Bobrovskikh\textsuperscript{1}
\hfill{}Pavel Karpyshev\textsuperscript{1}
\hfill{}Saveliy Galochkin\textsuperscript{1}
\hfill{}
\\\
\hfill{}Andrei-Timotei Ardelean\textsuperscript{1}
\hfill{}Arseniy Bozhenko\textsuperscript{1}
\hfill{}Ekaterina Karmanova\textsuperscript{1}
\hfill{}Pavel Kopanev\textsuperscript{1}
\hfill{}
\\\
\hfill{}Yaroslav Labutin-Rymsho\textsuperscript{3}
\hfill{}Ruslan Rakhimov\textsuperscript{1}
\hfill{}Aleksandr Safin\textsuperscript{1}
\hfill{}Valerii Serpiva\textsuperscript{1}
\hfill{}
\\\
\hfill{}Alexey Artemov\textsuperscript{4*}
\hfill{}Evgeny Burnaev\textsuperscript{1,2}
\hfill{}Dzmitry Tsetserukou\textsuperscript{1}
\hfill{}Denis Zorin\textsuperscript{5}
\hfill{}
\medskip\\\
\hfill{}\textsuperscript{1}Skolkovo Institute of Science and Technology
\hfill{}\textsuperscript{2}Artificial Intelligence Research Institute
\hfill{}
\\\
\hfill{}\textsuperscript{3}Moscow Engineering Physics Institute
\hfill{}\textsuperscript{4}Technical University of Munich
\hfill{}\textsuperscript{5}New York University
\hfill{}
}

\def\vec#1{\bm{#1}}
\def\ten#1{#1}

\def\parens#1{\left(#1\right)}
\def\braces#1{\left\{#1\right\}}
\def\brackets#1{\left[#1\right]}



\newcommand{\Kill}[1]{}
\definecolor{darkred}{HTML}{990000}
\iftoggle{nocomments}{
    \newcommand{\DZ}[1]{\ignorespaces}
    \newcommand{\EB}[1]{\ignorespaces}
    \newcommand{\OV}[1]{\ignorespaces}
    \newcommand{\LA}[1]{\ignorespaces}
    \newcommand{\AS}[1]{\ignorespaces}
    \newcommand{\todo}[1]{\ignorespaces}
    \newcommand{\RW}[1]{\ignorespaces}
}{
    \newcommand{\DZ}[1]{{\color{ForestGreen}DZ: #1}}
    \newcommand{\EB}[1]{{\color{magenta}EB: #1}}
    \newcommand{\OV}[1]{{\color{orange}OV: #1}}
    \newcommand{\LA}[1]{{\color{Plum}AA: #1}}
    \newcommand{\AS}[1]{{\color{blue}AS: #1}}
    \newcommand{\todo}[1]{{\color{red}TODO: #1}}
    \newcommand{\RW}[1]{{\color{darkred}#1}}
}
\newcommand{\notes}[1]{\ignorespaces}

\def\vs.{vs.\spacefactor=\the\sfcode`\v}
\def\etc.{etc.\spacefactor=\the\sfcode`\c}

\DeclareGraphicsExtensions{.eps,.pdf,.jpg,.png}
\iftoggle{lrgraphics}{
    \graphicspath{{src/figures/lr/}{src/figures/}}
}{
    \graphicspath{{src/figures/}}
}
\makeatletter
\iftoggle{nographics}{
	\LetLtxMacro{\includegraphics@orig}{\includegraphics}
	\RenewDocumentCommand{\includegraphics}{ s O{} m }{%
		{\setlength{\fboxsep}{0pt}%
		 \colorbox{lightgray}{\phantom{\IfBooleanTF{#1}{\includegraphics@orig*}{\includegraphics@orig}[#2]{#3}}}%
		}%
	}
}{}
\makeatother

\newcommand{\picbox}[2]{\fbox{\parbox{#1}{~\vspace{#2}}}}

    \iftoggle{separatesupp}{

    }{
\twocolumn[{%
\renewcommand\twocolumn[1][]{#1}%
\maketitle
\vspace{-10mm}
\begin{center}
\includegraphics[width=\textwidth]{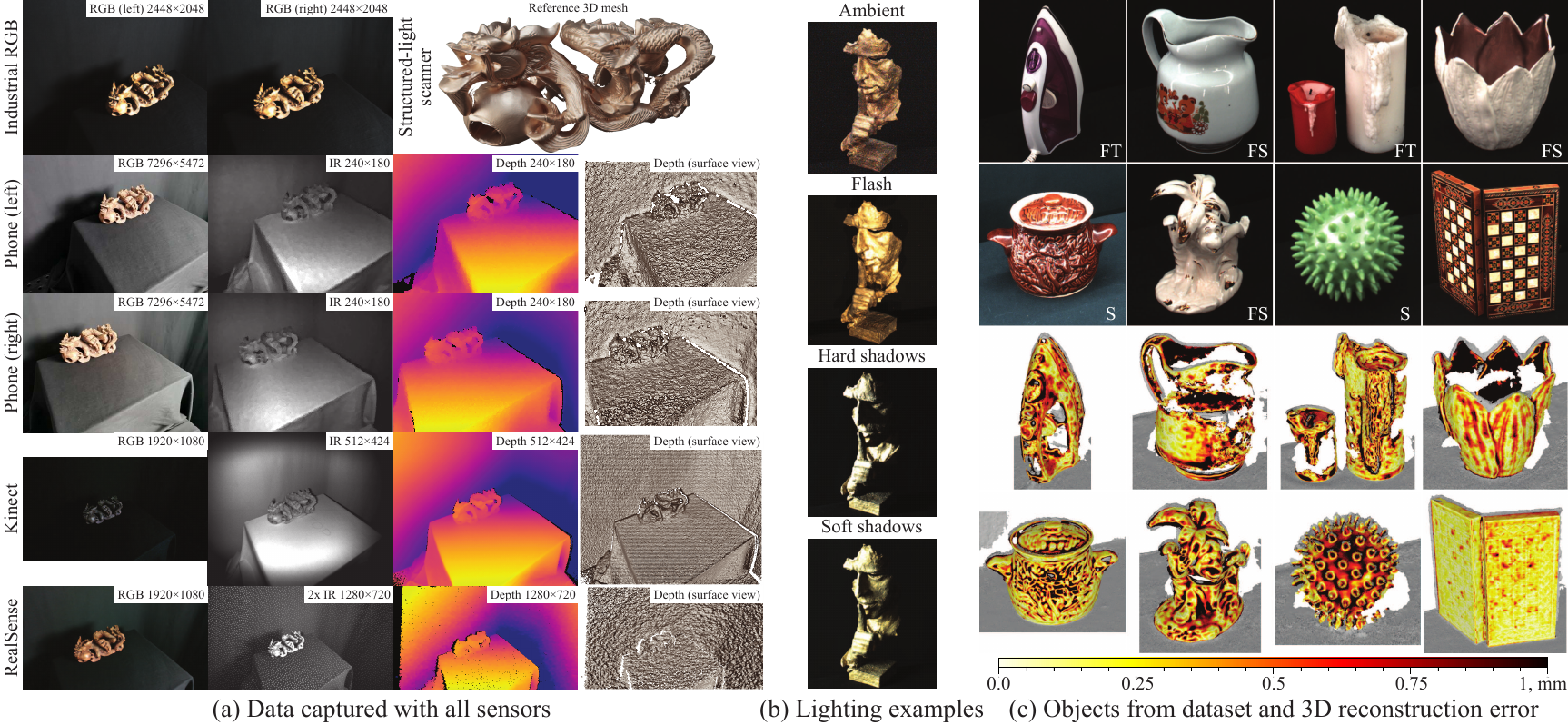}
\captionof{figure}{
Our dataset contains RGB-D data captured (a) with 7 different devices, (b) under various lightning conditions.
(c) We focus on materials challenging for 3D reconstruction algorithms,
such as featureless (F), highly specular with sharp reflections (S), or translucent (T),
as illustrated with reconstructions produced by state-of-the-art algorithms (compare with an \textquote{easy} object on the bottom right).
}
\label{fig:fig_teaser}
\end{center}
}]

\def\endabstract{}
\begin{abstract}
We present a new multi-sensor dataset for multi-view 3D surface reconstruction.
It includes registered RGB and depth data from sensors of different resolutions and modalities:
smartphones, Intel RealSense, Microsoft Kinect, industrial cameras, and structured-light scanner.
The scenes are selected to emphasize a diverse set of material properties challenging for existing algorithms.
We provide around 1.4 million images of 107 different scenes acquired from 100 viewing directions under 14 lighting conditions.
We expect our dataset will be useful for evaluation and training of 3D reconstruction algorithms and for related tasks.
The dataset is available at \href{http://skoltech3d.appliedai.tech}{\texttt{skoltech3d.appliedai.tech}}.
\end{abstract}

\renewcommand*{\thefootnote}{\fnsymbol{footnote}}
\footnotetext[1]{Work partially performed while at Skolkovo Institute of Science and Technology.}
\renewcommand*{\thefootnote}{\arabic{footnote}}
\setcounter{footnote}{0}

\section{Introduction}
\label{sec:intro}
Sensor data used in 3D reconstruction range from highly specialized and expensive CT, laser, and structured-light scanners
to video from commodity cameras and depth sensors;
computational 3D reconstruction methods are typically tailored to a specific type of sensors.
Yet, even commodity hardware increasingly provides multi-sensor data:
for example, many recent phones have multiple RGB cameras as well as lower resolution depth sensors.
Using data from different sensors, RGB-D data in particular, has the potential to considerably improve the quality of 3D reconstruction.
For example, multi-view stereo algorithms
(\eg,~\cite{schoenberger2016mvs-colmap,yao2018mvsnet})
produce high-quality 3D geometry from RGB data, but may miss featureless surfaces;
supplementing RGB images with depth sensor data makes it possible to have more complete reconstructions.
Conversely, commodity depth sensors often lack resolution provided by RGB cameras.

Learning-based techniques substantially simplify the challenging task of combining data fom multiple sensors.
However, learning methods require suitable data for training.
Our dataset aims to complement existing ones
(\eg,~\cite{DTU:jensen2014large,BigBIRD:singh2014bigbird,ScanNet:dai2017scannet,ETH3D:schops2017multi,TNT:knapitsch2017tanks,BlendedMVS:yao2020blendedmvs}),
as discussed in \cref{sec:related}, most importantly,
by providing multi-sensor data and high-accuracy ground truth for objects with challenging reflection properties.

The structure of our dataset is expected to benefit research on 3D reconstruction in several ways.
\begin{itemize}
    \item\emph{Multi-sensor data.}
We provide  data from \emph{seven} different devices with high-quality alignment,
including low-resolution depth data from commodity sensors (phones, Kinect, RealSense),
high-resolution geometry data from a structured-light scanner, and  RGB data at different resolutions and from different cameras.
This enables supervised learning for reconstruction methods relying on different combinations of sensor data,
in particular, increasingly common combination of high-resolution RGB with low-resolution depth data.
In addition, multi-sensor data simplifies comparison of methods relying on different sensor types (RGB, depth, and RGB-D).

    \item \emph{Lighting and pose variability.}
We chose to focus on a setup with controlled (but variable) lighting and a fixed set of camera positions for all scenes,
to enable high-quality alignment of data from multiple sensors, and systematic comparisons of algorithm sensitivity to various factors.
We aimed to make the dataset large enough  (1.39\,M images of different modalities in total) to support training machine learning algorithms,
and provide systematic variability in camera poses (100 per object),
lighting (14 lighting setups, including \textquote{hard} and diffuse lighting, flash, and backlighting, as illustrated in~\cref{fig:fig_teaser}\figcolor{b}),
and reflection properties these algorithms need.

    \item \emph{Object selection.}
Among 107 objects in our dataset, we include primarily objects
that may present challenges to existing algorithms mentioned above (see examples in \cref{fig:fig_teaser}\figcolor{c});
we made special effort to improve quality of 3D high-resolution structured-light data for these objects.
\end{itemize}

Our focus is on RGB and depth data for individual objects in laboratory setting, similar to the DTU dataset~\cite{DTU:jensen2014large},
rather than on complex scenes with natural lighting, as in Tanks and Temples~\cite{TNT:knapitsch2017tanks} or ETH3D~\cite{ETH3D:schops2017multi}.
This provides means for systematic exploration and isolation of different factors  contributing to strengths and weaknesses of different algorithms,
and complements more holistic evaluation and training data provided by datasets with complex scenes.

\begin{table}[t]
\centering
\begin{tikzpicture}
    \node[anchor=south west,inner sep=0] (image) at (0,0) {\includegraphics[width=\columnwidth]{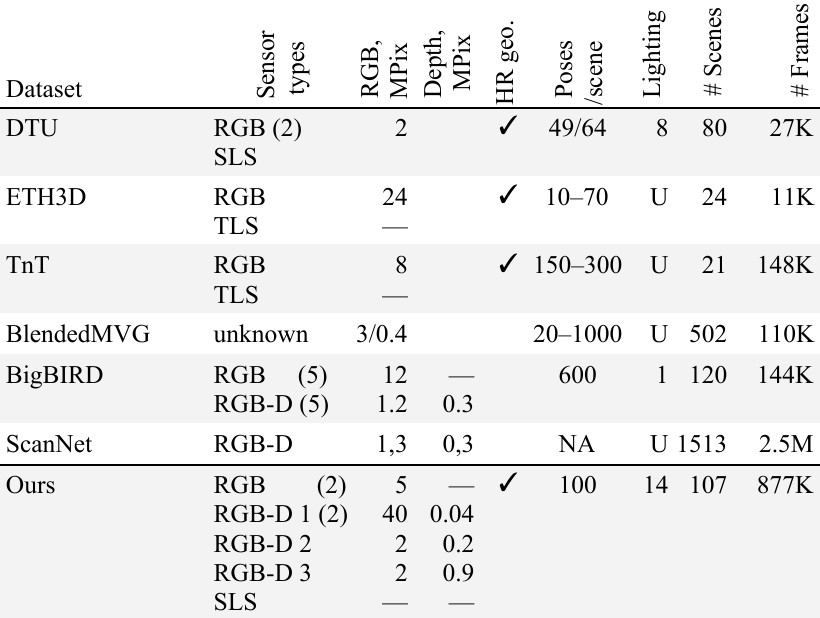}};
    \begin{scope}[shift=(image.north west), x={(image.north east)},y={(image.south west)}]
        \footnotesize
        \node at (.101, .21) {\cite{DTU:jensen2014large}};
        \node at (.134, .322) {\cite{ETH3D:schops2017multi}};
        \node at (.092, .433) {\cite{TNT:knapitsch2017tanks}};
        \node at (.213, .543) {\cite{BlendedMVS:yao2020blendedmvs}};
        \node at (.155, .61) {\cite{BigBIRD:singh2014bigbird}};
        \node at (.145, .721) {\cite{ScanNet:dai2017scannet}};
    \end{scope}
\end{tikzpicture}
\caption{
\textbf{Comparison of our dataset to the most widely used related datasets.}
U indicates uncontrolled lighting;
frames are counted per sensor, \ie., all data from an RGB-D sensor are counted as a single frame.
The number of separate images acquired may be considerably larger (1.4\,M for our dataset).
All scenes, from both training and testing sets, were counted.
}
\label{tab:tbl_dataset_comparison}
\end{table}

\section{Related work}
\label{sec:related}
Many datasets for tasks related to 3D reconstruction were developed
(see, for example, a survey of datasets related to simultaneous localization and mapping (SLAM)~\cite{liu2021simultaneous});
we only discuss datasets most closely related to ours.
A summary of comparisons to key datasets from previous work is shown in~\cref{tab:tbl_dataset_comparison}.

\paragraph{RGB datasets with high-resolution 3D ground truth.}
A number of datasets are designed for multi-view stereo (MVS) methods,
such as PatchMatch-based~\cite{Barnes2009PatchMatch,schoenberger2016mvs-colmap,xu2019ACMM,xu2020ACMP},
learning-based~\cite{yao2018mvsnet,Leroy_2018_ECCV,gu2020casmvsnet,zhang2020vismvsnet,ma2021epp-mvsnet,wang2021patchmatchnet,wei2021aa-rmvsnet,VolumeFusion:Choe_2021_ICCV},
or hybrid methods~\cite{DeFuSR:Donne_2019_CVPR,Kuhn2020DeepCMVS}.
These datasets are also used for evaluation of
methods reconstructing an implicit surface representation encoded by a neural network from a set of RGB
images~\cite{DVR:niemeyer2020differentiable,IDR:yariv2020multiview,wang2021neus,oechsle2021unisurf},
and in the novel view synthesis task~\cite{mildenhall2020nerf,aliev2020neural,wang2021ibrnet,riegler2021stable,rakhimov2022npbgpp}.

Datasets from this category typically include high-resolution RGB,
either photo~\cite{DTU:jensen2014large,aanaes2016large,ETH3D:schops2017multi,BlendedMVS:yao2020blendedmvs}
or video~\cite{TNT:knapitsch2017tanks},
and high-resolution 3D ground truth obtained with
a structured-light scanner (SLS)~\cite{aanaes2016large} or a terrestrial laser scanner (TLS)~\cite{ETH3D:schops2017multi,TNT:knapitsch2017tanks}.
The Middlebury datasets~\cite{MiddleburyMVS:seitz2006comparison,Middlebury14:Scharstein2014,scharstein2014high}
focus on two-frame stereo, providing accurate ground truth for disparity in addition to RGB.

In this group, the DTU dataset~\cite{DTU:jensen2014large} with controlled lighting and high-resolution ground truth
is most often used for training learning-based MVS methods.
Most other MVS datasets, while containing some images of isolated objects, focus on complete scenes,
often collected with hand-held, freely positioned cameras~\cite{strecha2008benchmarking,ETH3D:schops2017multi,TNT:knapitsch2017tanks}.

Compared to previously developed datasets with high-resolution 3D scanner data,
we provide the largest number of sensors, objects, and lighting conditions, and the most challenging objects.

\paragraph{Datasets with low-resolution depth.}
Datasets designed for tasks like SLAM, object recognition and segmentation
are often collected using low-resolution depth sensors like Microsoft Kinect
or Intel RealSense~\cite{NYUv2:Silberman2012,BigBIRD:singh2014bigbird,SUN_RGBD:song2015sun,Redwood:Park2017,ScanNet:dai2017scannet,chang2017matterport3d};
some of these datasets combine high-resolution RGB with low-resolution depth
(\eg,~\cite{BigBIRD:singh2014bigbird,ScanNet:dai2017scannet})
but do not provide high-resolution ground truth for depth.
A notable exception is CoRBS dataset~\cite{CoRBS:wasenmueller2016corbs}, but it contains only four scenes.
The degree of control over camera position and lighting in these datasets varies
from complete~\cite{TUM:Cremers-Kolev-pami11,BigBIRD:singh2014bigbird} to none (\eg,~\cite{Redwood:Park2017}).

These datasets are often used for qualitative evaluation of non-learning-based
depth fusion methods~\cite{izadi_kinectfusion:2011,whelan_elasticfusion:2016,dai_bundlefusion:2017,schops_surfelmeshing:2020}
which produce voxel- or surfel-based surface representations from depth maps;
they are also used to train learning-based methods which produce voxel-based surface representation
from RGB~\cite{NeuralRecon:Sun2021CVPR,murez2020atlas}, depth~\cite{SG-NN:Dai_2020_CVPR,dai2018scancomplete}, or RGB-D~\cite{SPSG:Dai_2021_CVPR} data.
These methods are likely to benefit from including our high-resolution depth data in training.

\begin{figure}[t]
\centerline{\includegraphics[width=\columnwidth]{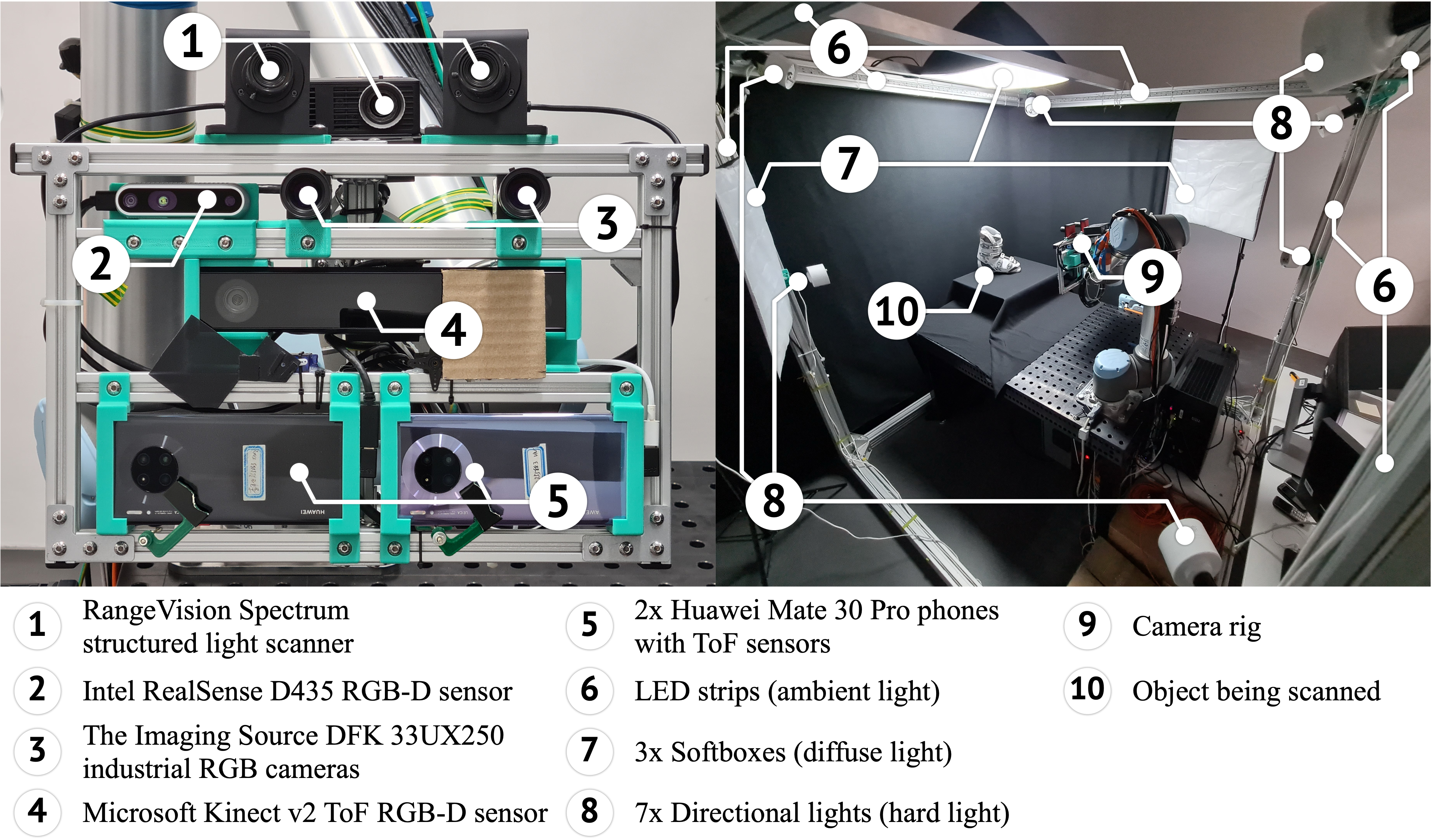}}
\caption{
\textbf{Our acquisition setup} (view in zoom).
We included a diverse set of seven commonly used RGB and RGB-D sensors, mounting them on a shared metal rig to aid data alignment.
We constructed a metal frame surrounding the scanning area and installed various light sources to provide 14 lighting setups.
}
\label{fig:rig_and_room}
\end{figure}

\paragraph{Synthetic datasets}
ShapeNet~\cite{shapenet2015} and ModelNet~\cite{ModelNet:CVPR15_Wu} are often used to train learning-based depth fusion methods,
\eg,~\cite{weder_routedfusion:2020,NeuralFusion:Weder_2021_CVPR,DI-Fusion:Huang_2021_CVPR,DeepSurfels:Mihajlovic_2021_CVPR}.
Synthetic ICL-NUIM SLAM benchmark~\cite{ICL-NUIM:Handa_2014} is used for evaluation,
for example, in~\cite{whelan_elasticfusion:2016,dai_bundlefusion:2017,schops_surfelmeshing:2020}.
Such approach is limited by the differences between real-world and synthetic data.

Synthetic benchmarks~\cite{berger2013benchmark,ley2016syb3r,koch2021hardware} allow to generate large training sets easily
via simulation of the acquisition process.
However, real-world data is still required to faithfully model sensors, train generators, and test trained algorithms.
Our dataset can be used for these purposes.

\paragraph{Datasets with multiple depth resolutions.}
Recently proposed RGB-D-D~\cite{RGB-D-D:He_2021_CVPR} and {ARK}itScenes~\cite{dehghan2021arkitscenes} datasets
make a step in a similar direction to ours,
pairing low-resolution depth data acquired by smartphones with medium resolution (0.3\,MPix) time-of-flight data
and high-resolution laser scans, respectively.

Our dataset contains inputs from multiple depth sensors of low and high resolutions along with associated registered higher resolution RGB images,
providing a framework for evaluating and training both depth fusion and RGB-D fusion algorithms
previously trained on synthetic data, as well as developing new ones.
Having multiple depth resolutions also supports applications such as
depth map super-resolution~\cite{MSG-Net:Hui2016,MS-PFL:Chuhua2020,CA-IRL_SR:Song_2020,voynov2019perceptual}
and depth map completion~\cite{DC:Zhang2018,RecScanNet:Jeon2018}.

Similarly to our dataset, a concurrent work~\cite{shrestha2022real} complements RGB-D sequences from Intel RealSense
with registered ground-truth captured by structured-light scanning.
We provide registered depth images from common sensors at three levels of accuracy,
and a controlled lighting variation.

\section{Dataset}
\label{sec:dataset}

\begin{table}[t]
\centering
\includegraphics[width=\columnwidth]{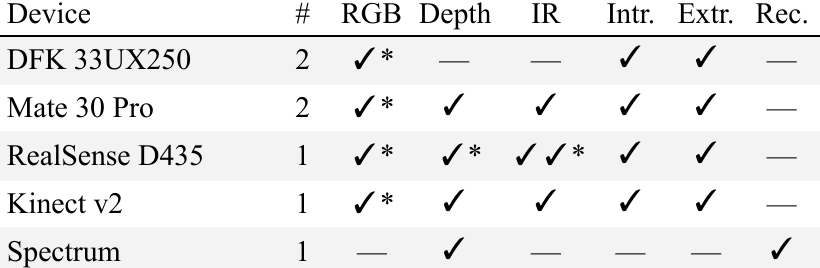}
\caption{
\textbf{Composition of our dataset.}
We provide RGB, depth, and IR images, intrinsic (Intr.) and extrinsic (Extr.) calibration parameters,
and a reference mesh reconstruction (Rec.).
The data marked with * is captured per lighting setup.
}
\label{tab:tbl_data_composition}
\end{table}

\subsection{Overview}
\label{dataset:overview}

Our dataset consists of 107 scenes with a single everyday object or a small group of objects on a black background,
see examples in \cref{fig:fig_teaser} and a complete list of scenes in the supplementary material.

We collected the dataset using multiple sensors mounted on Universal Robots UR10 robotic arm
with 6 degrees of freedom and sub-millimeter position repeatability.
We used the sensors shown in \cref{fig:rig_and_room} on the left:
(1) RangeVision Spectrum structured-light scanner (SLS),
(2) two The Imaging Source DFK 33UX250 industrial RGB cameras,
(3) two Huawei Mate 30 Pro phones with time-of-flight (ToF) depth sensors,
(4) Intel RealSense D435 stereo RGB-D camera,
and (5) Microsoft Kinect v2 ToF RGB-D camera.

We surrounded the scanning area with a metal frame to which we attached the light sources,
shown in \cref{fig:rig_and_room} on the right:
seven directional lights, three diffuse soft-boxes, and LED strips which imitate ambient light.
We also used flashlights on the phones as the light source moving with the camera.
To prevent cross-talk between depth sensors we added external shutters
that close the infrared (IR) projector of one sensor while the others are imaging.

For each scene, we moved the camera rig through 100 positions on a sphere with a radius of 70~cm around the object,
using the same trajectory for all scenes,
and collected the data using 14 lighting setups.
For each device, except the SLS, we collected raw RGB, depth, and IR images,
including both left and right IR images for RealSense.
In total, we collected 15 raw images per scene, camera position, and lighting setup: 6 RGB, 5 IR, and 4 depth images,
as illustrated in \cref{tab:tbl_data_composition,fig:fig_teaser}\figcolor{a}.
As the IR and depth data from ToF sensors of the phones and Kinect is unaffected by the lighting conditions,
we captured this data once per camera position.
For the SLS we collected partial scans from 27 positions and combined them into a single scan, as we describe further.

In our dataset, we included a large number of objects with challenging and varied surface material properties,
as shown in~\cref{tab:dataset_feature_stats}.
A set of qualitative labels corresponding to the key surface reflection parameters were assigned to each object
based on visual estimation of these parameters.
For example, \emph{Specularity} represents the ratio of specular to diffuse reflectance for one of the dominant materials of the object,
and \emph{Reflection sharpness} characterizes how sharp the reflectance function peak is.
These labels and their relation to performance of 3D reconstruction methods are discussed in the supplementary material.

\subsection{Data acquisition and post-processing}
\label{dataset:acquisition}

Here, we summarize the main steps of our data acquisition,
including selection of camera settings, camera calibration, and preparation of objects for scanning,
and then describe our post-processing pipeline,
which employs new methods for high-accuracy data registration
and automatic generation of occlusion-valid reference depth images from SL scans.

\noindent \textbf{The data acquisition procedure}
for each scene consisted of the following steps:
\begin{enumerate}
\item We automatically selected optimal camera exposure and gain settings for the scene.
\item We performed a preliminary SL scan, and if it was incomplete or low quality due to surface reflection properties,
we applied a vanishing opaque matte coating to the object.

\item We scanned the object with the SLS from 27 viewpoints.

\item For coated objects, we accelerated the coating sublimation with hot air while keeping the object stationary.
We then collected additional validation SL scans from five viewpoints to verify that the object was not deformed during coating removal.

\item We acquired RGB and low-resolution depth sensor data.
\end{enumerate}

\noindent \textbf{Sensor exposure/gain selection}
is critical for obtaining useful data:
uniform settings result in frequent over- or underexposure due to variations of object surface properties.
Hardware auto-exposure, being biased by the black background,
proved to be inadequate for our setup and produced too high exposure/gain.
Instead, we designed an auto-exposure algorithm inspired by~\cite{shin2019camera,shim2018gradient},
which we used to find optimal camera settings for each scene, lighting, and sensor individually.

To find the \emph{minimal noise} camera settings,
we extracted the foreground mask of the scene from the images with and without the object using the method of~\cite{lin2021real},
and then maximized the Shannon entropy of the foreground image \wrt~the exposure value while keeping the gain at minimum.
We then additionally maximized the entropy \wrt~the gain value while keeping the found exposure value,
to prevent the underexposure of very dark objects.
To find the \emph{real-time / high-noise} camera settings we swapped exposure and gain,
optimizing the gain first while keeping the exposure at 30\,FPS, and then optimizing the exposure while keeping the found gain.

We obtained \emph{minimal noise} camera settings for each lighting variant,
and \emph{real-time} settings only for the dim flash and ambient lighting.
We picked the settings for a single position of the rig per scene and used these settings for all positions.
For RealSense, we also picked the optimal power of the IR projector out of 12 options,
selecting the one leading to the lowest number of depth pixels missing a value.

\begin{table}[t]
    \centering
    \includegraphics[width=\columnwidth]{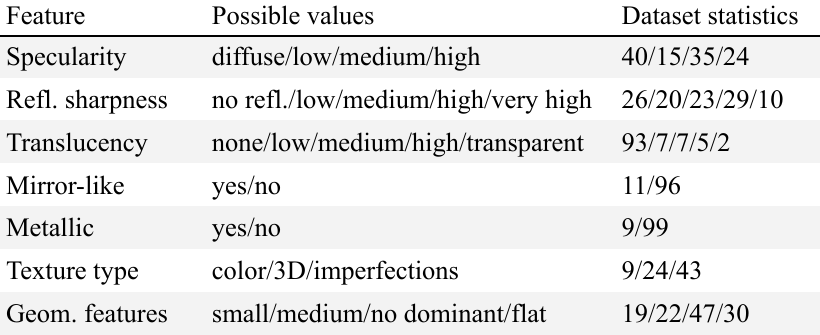}
    \caption{\textbf{Surface material properties in our dataset.}}
    \label{tab:dataset_feature_stats}
\end{table}

\begin{wrapfigure}{r}{0.47\columnwidth}
\vspace{-4pt}
\centerline{\includegraphics[width=.2\columnwidth]{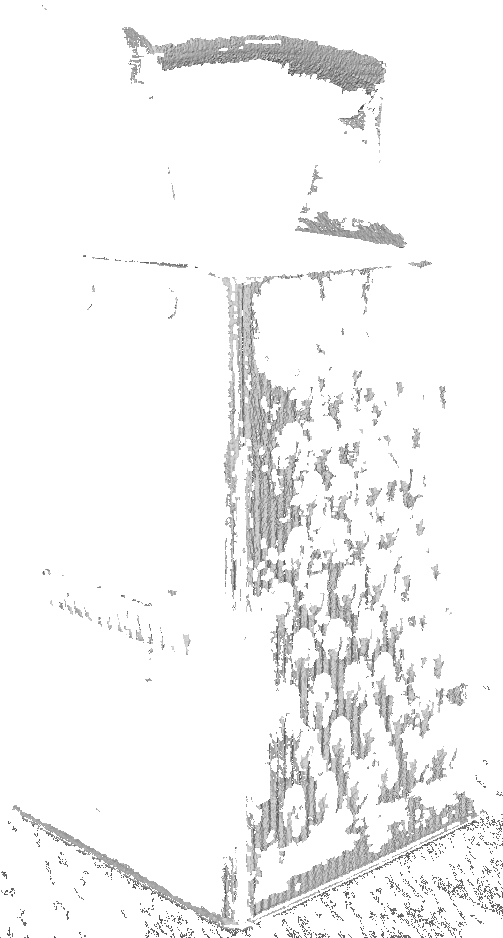}\includegraphics[width=.2\columnwidth]{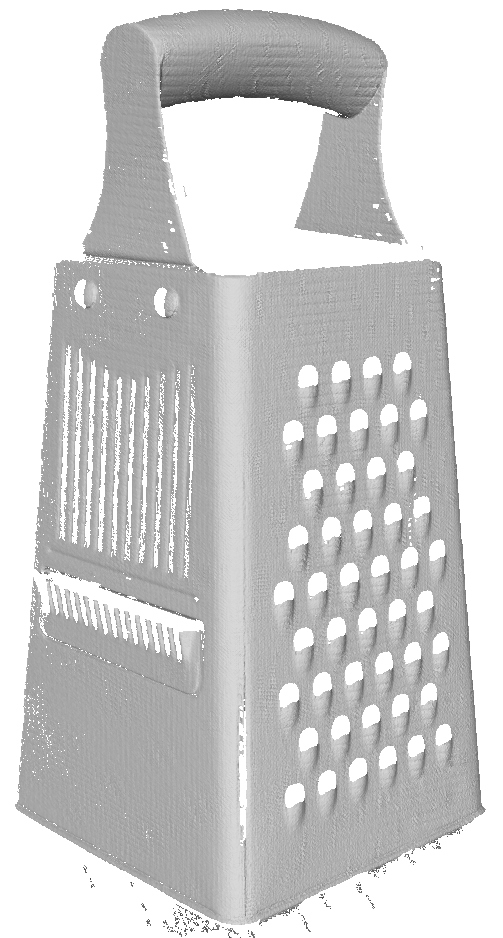}}
\caption{A partial scan without (left) and with coating (right).}
\label{fig:structlight_failure}
\vspace{-8pt}
\end{wrapfigure}

\noindent \textbf{Preparation of objects for 3D scanning.}
For our dataset
we picked the objects with surface material properties that challenge common sensors and reconstruction algorithms.
However, these properties often challenge the SLS too,
making it hard to obtain reliable high-resolution reference data.
To get the highest-quality SL scans we applied a temporary coating to about half of the scanned objects
(see~\cref{fig:structlight_failure}).

\noindent \textbf{Post-processing of structured-light scans.}
To simplify the use of the SL data for evaluation and training we merged raw partial scans into a single clean surface reconstruction.
For this, we globally aligned partial scans using a variant of the iterative closest point algorithm,
initialized with calibrated scanner poses.
We assessed the alignment quality by comparing the inter-scan distance to the scanner resolution
and in case of poor alignment, due to a lack of geometrical features on the object, re-scanned the object with attached 3D markers.
From the aligned scans, we reconstructed a surface mesh using screened Poisson Surface Reconstruction~\cite{kazhdan2013screened}
with the cell size of 0.3\,mm, corresponding to a conservative estimate of the scanner resolution.
Finally, we cleaned the surface keeping only the vertices close to the raw scans
and manually removing the scanning and reconstruction artifacts and the supporting stand.
We refer to the resulting surface as \emph{the SL scan}.

For the temporary coating of the scanned objects we used Aesub Blue scanning spray.
It sublimates from the surface at room temperature in a few hours,
which we reduced to 5--15\,minutes by slightly heating the object with a heat gun.
To detect potential object deformations, we aligned the five validation scans to the SL scan
and visually checked for distance variation indicating deformation.
If a deformation was localized only to an isolated part of the object (\eg, a power cord) we removed this part from the scan,
otherwise, we excluded the whole object from the dataset.

\noindent \textbf{Camera calibration.}
To measure the trajectories of the sensors and their intrinsic camera parameters
we used the calibration pipeline of~\cite{schops2020having}.
It is based on generic camera models known to result in a more accurate calibration than parametric camera models,
commonly used in benchmarks related to ours, including DTU and ETH3D.

Application of this calibration pipeline to our camera rig as is was numerically unstable
due to the large variability of sensor properties, such as field of view and resolution.
To eliminate instability we split the calibration procedure into several steps,
as we describe in the supplementary material.

For all RGB and depth sensors and the SLS, we obtained central generic camera models with several thousands parameters
and the trajectory of the sensor in the global coordinate system.
The mean calibration error for different sensors at different steps of the procedure was in the range 0.04--0.4\,px,
or approximately in the range 0.024--0.15\,mm.

We observed thermal effects to contribute noticeably to calibration results,
in particular, calibration errors drop by factors of 2--10 if the devices warm up after power-on to stable operating temperature.
To reduce thermal drift in our data we pre-warmed all devices for~1~hour
before calibration and at the beginning of each day of scanning.

\begin{figure}[t]
\centerline{\includegraphics[width=.5\columnwidth]{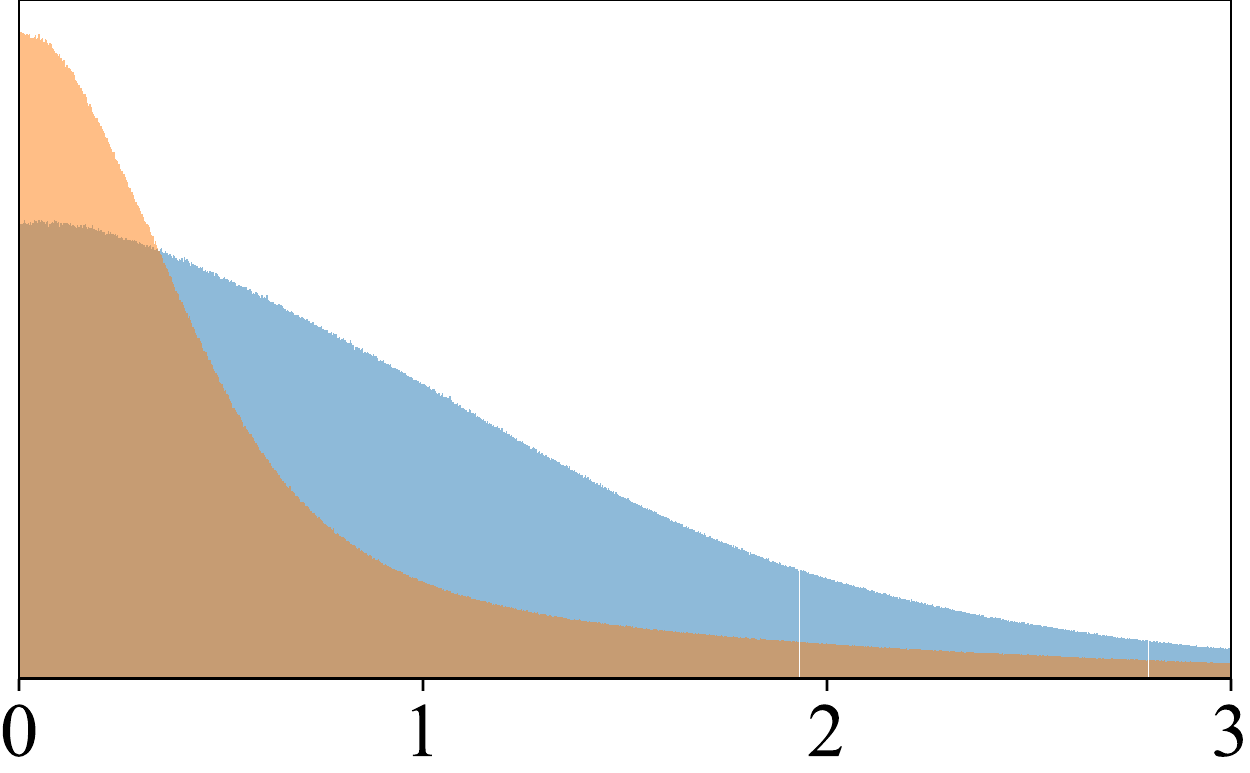}
            \includegraphics[width=.5\columnwidth]{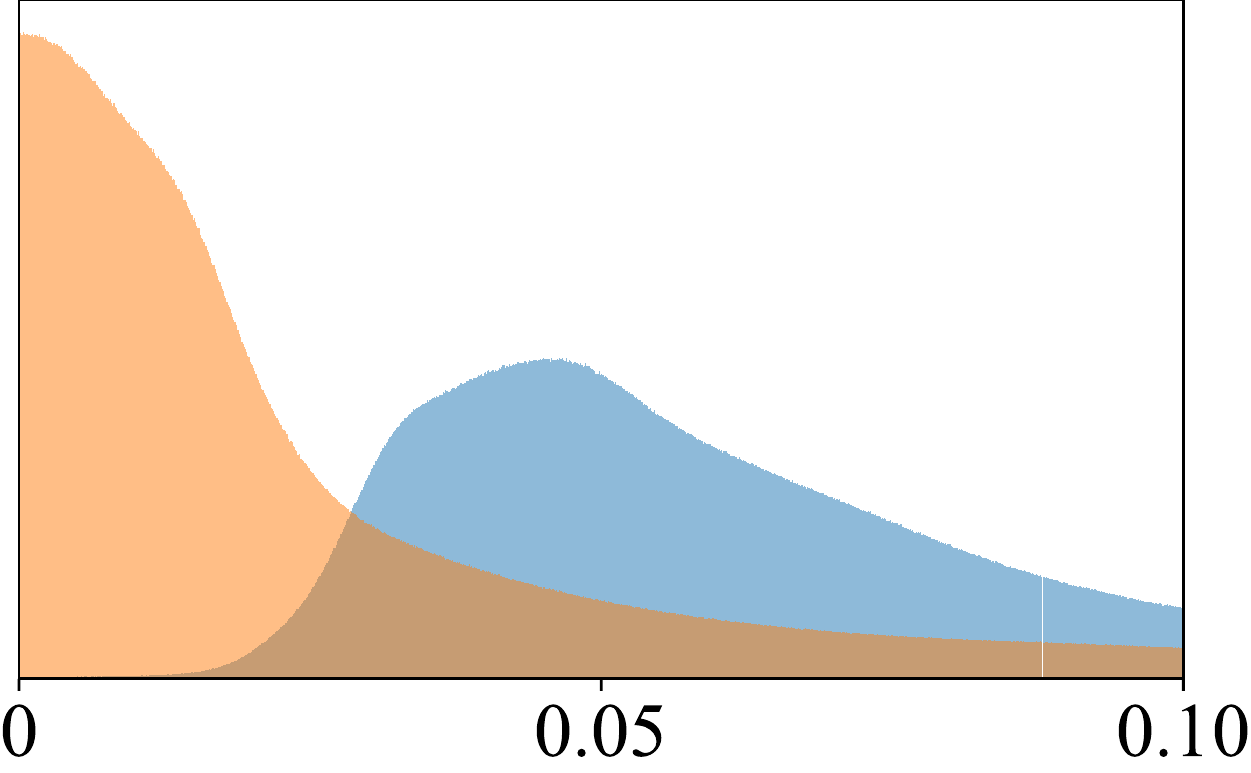}}
\caption{
\textbf{Calibration refinement results.}
Left: distribution of COLMAP reconstruction errors in mm before camera pose refinement (blue) and after (orange).
Right: distribution of relative Kinect depth errors \wrt.~SLS before correction (blue) and after (orange).
}
\label{fig:fig_cam_pose_and_sensor_depth_correction}
\end{figure}

\noindent \textbf{Refinement of camera poses.}
Limited rig stiffness and variations in positioning of the robotic arm lead to a noticeable deviation
of the camera trajectory during scanning from the calibrated trajectory, up to several pixels in image space.
We additionally refined individual sensor poses per scene \wrt~SL scan
using a new approach inspired by~\cite{corsini2009image,dellepiane2013global}.

The idea of these works is to optimize a photometric discrepancy between a sensor image
and a render of a reference texture on the scan to the image plane, with respect to the pose of the sensor.
In~\cite{corsini2009image}, geometric features of the scan, such as normals, are used as the reference texture,
while in~\cite{dellepiane2013global}, poses for multiple viewpoints are optimized simultaneously
and for each viewpoint the projections from images from all other viewpoints are used as the reference.
In both cases, the mutual information between intensity distributions of the sensor image and the render is used as the discrepancy measure,
and it is optimized using derivative-free NEWUOA method~\cite{powell2006newuoa}.
We used the same overall idea but changed both key components of the method:
the reference texture and the discrepancy measure.

The modification of the original method was required
as the previously proposed variants did not yield sufficiently accurate results for our data.
The variant of~\cite{corsini2009image}
assumes the presence of photometric similarity between the render of the scan surface and the photo of the real surface,
which is often not present, \eg, for a smooth surface with color texture,
and, importantly, relies on the match between the silhouettes of the rendered scan and the real object in the photo,
which is often violated as not all parts of the object visible in the photo may be scanned.
Simultaneous optimization of multiple viewpoints in~\cite{dellepiane2013global} often results in pose drift.

Our approach is to use the raw SLS camera images as the reference texture,
since the scan is constructed from these images and they are perfectly aligned.
Instead of mutual information between intensity distributions, which we found to be often unstable,
we optimized the smooth \(L_1\) distance in deep CNN feature space.
Inspired by~\cite{lindenberger2021pixel},
for feature extraction we used a model pre-trained specifically for feature matching~\cite{germain2020s2dnet}.

For some camera poses and sensors, the direct alignment to the SLS images was unstable,
as these images were captured under SLS illumination, which differs significantly from the illumination used for the other cameras.
To further stabilize the alignment procedure for the whole dataset, we split it into three steps.
First, we aligned, directly to the SLS reference texture, a single image from an industrial RGB camera,
for a viewpoint selected for each scene individually to maximize the alignment tightness.
Then, we aligned the images from the industrial RGB camera for the remaining viewpoints successively,
using the already aligned images as the reference.
Finally, for all the other sensors, we aligned each image individually
to the image from the industrial RGB camera captured at the same position of the rig.

Using this sequential sensor alignment procedure we obtained subpixel alignment accuracy in most cases.
In~\cref{fig:fig_cam_pose_and_sensor_depth_correction} left,
we show that refinement of camera poses reduces the error of a 3D reconstruction method.

\noindent \textbf{Refinement of depth camera calibration.}
The lower quality depth sensors were calibrated by the manufacturers
using calibration systems different from the one we used for the SLS and the camera trajectories.
This leads to a misalignment between the lower quality depth data and the SL scan,
which can be represented as a composition of a 3D rigid transform, scaling and a non-linear
warping~\cite{herrera2012joint,teichman2013unsupervised,zhou2014simultaneous,zeisl2016structure}.

To reduce the misalignment, we used a correction model
\(d_{\text{undist}} = S(u, v) S(d_{\text{raw}})\),
which transforms the raw depth measurement \(d_{\text{raw}}\) at pixel $(u, v)$ in the depth image
to the depth value in the 3D space of the SL scan,
using the cubic basis splines $S(u, v)$ and $S(d)$ defined on regular grids.
We trained the model on depth images of a calibration pattern,
and as the reference used the calibration data from the same system that we used for camera calibration.

In~\cref{fig:fig_cam_pose_and_sensor_depth_correction} right,
we show that refinement of depth camera calibration reduces the error of a low quality sensor depth \wrt~the SL scan.

\begin{figure}[t]
\centerline{\includegraphics[width=.332\columnwidth]{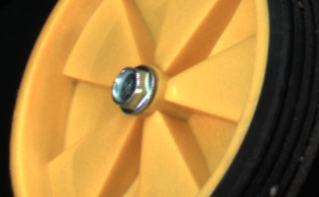}
            \includegraphics[width=.332\columnwidth]{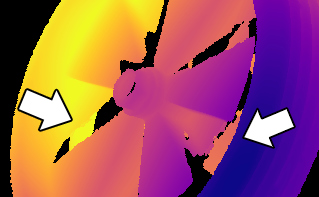}
            \includegraphics[width=.332\columnwidth]{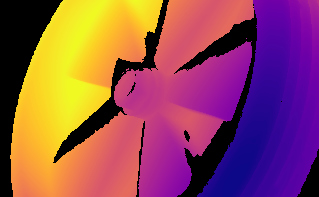}}
\caption{
\textbf{Rendering SL depth with occluding surface.}
Although the real surface of the object is solid (left),
its SL scan is incomplete, so the background parts of the object bleed through in the rendered depth map (middle).
Occluding surface helps to discard incorrect depth values (right).
}
\label{fig:fig_vis_occlusion_result}
\end{figure}

\noindent \textbf{Rendering SL scans. Occluding surface.}
In many tasks that we target, the ground-truth 3D data is needed in the form of depth maps.
At the same time, the SL scanning often does not capture the complete object due to its geometric limitations,
so a direct rendering of a depth map from the SL scan is likely to produce incorrect large depth values at some points
because closer parts of the surface are absent from the scan,
as illustrated in~\cref{fig:fig_vis_occlusion_result}~middle.

Previous works have addressed this problem by discarding depth values
either inconsistent with an MVS reconstruction from RGB images~\cite{luo2020attention},
or occluded by a complete surface reconstructed from the scan and adjusted manually~\cite{ETH3D:schops2017multi}.
We follow the latter, more reliable approach, but propose a fully automatic method for building the occluding surface.

An \emph{occluding surface} for an object is any surface fully enclosing the object.
Our goal is to construct an occluding surface tight to the real surface of the object,
using the information about visibility of the real surface from the viewpoint of the SLS camera during scanning.
The depth values rendered from the SL scan
that differ from the values rendered from such an occluding surface by more than a certain threshold
can then be identified as spurious and removed from the generated depth maps.

{
\def\cone{\mathrm{Cone}(v_i, S_i)}

To find the occluding surface we assume that the convex hull of all partial SL scans \(C\)
encloses the part of the object that we are going to render.
Conceptually, we find the occluding surface as
\(C \setminus \cup_i \cone\),
where the cone \(\cone\) encloses all the points between the surface of the partial scan \(S_i\)
and the respective viewpoint of the SLS camera \(v_i\).
Intuitively, we carve the free space between the scanner and the object out of the convex hull,
for all positions of the scanner.

The direct computation of such an occluding surface by a sequence of boolean operations on closed meshes
is extremely expensive as the number of triangles for each mesh can be very large.
Instead, we compute an accurate approximation by sampling all surfaces involved
(we used the sampling distance of 0.1\,mm)
and keeping only the points that are in the interior of the convex hull and in the exterior of all cones,
which we check using depth testing.
We then reconstruct the occluding surface from these samples using screened Poisson Surface Reconstruction.
}

The occluding surface obtained with this method can be used to generate accurate, occlusion-valid depth images from SL data
as shown in Figure~\ref{fig:fig_vis_occlusion_result} right.

\section{Experimental evaluation}
\label{sec:experiments}

We demonstrate the challenges of our dataset for RGB-based 3D reconstruction methods,
and additionally, demonstrate one of the uses of our dataset
as a benchmark to compare methods of 3D reconstruction from different modalities.
For this, we tested
5 methods which reconstruct the surface only from RGB data,
3 methods which use only depth data,
and one method which uses both modalities.

\noindent \textbf{Methods.}
COLMAP~\cite{schoenberger2016sfm-colmap,schoenberger2016mvs-colmap} is an RGB-to-3D reconstruction pipeline.
It implements a non-learning-based multi-view stereo (MVS) method: from multi-view RGB images, the depth maps are estimated per-view
and then fused into a point cloud representing the reconstructed 3D surface.
COLMAP is commonly evaluated on benchmarks similar to ours, so we include it as a baseline.
ACMP~\cite{xu2020ACMP} is a non-learning, PatchMatch-based MVS method with planar prior,
which shows a strong performance on benchmarks such as Tanks and Temples (TnT).
VisMVSNet~\cite{zhang2020vismvsnet} and UniMVSNet~\cite{peng2022rethinking} are learning, plane-sweeping-based MVS methods
with top performance on TnT benchmark.
NeuS~\cite{wang2021neus} is a method which reconstructs the surface as a signed distance function (SDF)
represented with a neural network which is fitted directly to RGB images via differentiable rendering.
TSDF Fusion~\cite{curless_volumetric_1996,izadi_kinectfusion:2011} is a classical non-learning-based depth-fusion approach.
It reconstructs a truncated SDF (TSDF) of a surface on a voxel grid via iterative integration of depth maps.
RoutedFusion~\cite{weder_routedfusion:2020} is a learning-based extension, which performs depth map integration using neural networks.
SurfelMeshing~\cite{schops_surfelmeshing:2020} is a non-learning-based depth-fusion method that reconstructs the surface
using a surfel cloud as an intermediate representation.
Neural RGB-D surface reconstruction~\cite{azinovic2022neural} is a recent method which,
similarly to NeuS, reconstructs the TSDF of a surface represented with a neural network,
but in contrast to NeuS, fits the network to RGB \textit{and depth} images.

\noindent \textbf{Data.}
We tested the methods under the most favorable conditions available in our dataset.
We obtained the reconstructions using full-resolution images from all 100 viewpoints for each scene,
and the ground-truth camera poses and intrinsic camera models.
For each data modality we used the highest-quality option:
the photos from one of the industrial RGB cameras with the minimal noise settings,
taken under ambient light to rule out the effects of lighting,
and the depth maps from the ToF sensor of Kinect, of a higher resolution than of the phones, and more stable than RealSense.

For the learning-based methods we tested the models trained on prior datasets:
DTU~\cite{DTU:jensen2014large} and BlendedMVS~\cite{BlendedMVS:yao2020blendedmvs} for MVSNets,
and ModelNet~\cite{ModelNet:CVPR15_Wu} for RoutedFusion.
We provide more details in the supplementary material.

\noindent \textbf{Measures.}
We evaluated the reconstructions \wrt~the full SL scans using Precision, Recall, and F-score quality measures,
similarly to the prior benchmarks~\cite{ETH3D:schops2017multi,TNT:knapitsch2017tanks}.
Precision measures the reconstruction accuracy:
we calculated it as the percentage of the reconstruction points closer to the reference than a certain distance threshold.
Recall measures the reconstruction completeness:
we calculated it as the percentage of the \textit{reference} points closer to the reconstruction than a threshold.
F-score is the harmonic mean of these two numbers,
which, to be high, requires the reconstruction to be both accurate and complete.

For a careful calculation of Precision and Recall two problems have to be considered.
First, both the reconstruction and the reference may have varying point densities,
which will cause uneven contribution of different parts of the surface to the value of the measure.
Second, the reference SL scans are incomplete, so the distance from some reconstruction points to the SL scan
does not represent the distance to the real surface of the object,
specifically, if the point lies near the missing part of the surface.
We addressed these problems similarly to~\cite{ETH3D:schops2017multi}, extending the approach using our new occluding surface.
Specifically, we calculated each measure in small cells of 3D space and then took the average as the final value,
and used only the points for which the distance to the real surface of the object is certain.
We describe this in more detail in the supplementary material.

\noindent \textbf{Results.}
In~\cref{fig:fig_measures_vs_scenes_n} we show the average performance on our dataset for the tested methods.
Each curve shows the number of scenes that the respective method reconstructs with a better value of the measure
than the value on the X~axis.
We tested Neural RGB-D surface reconstruction only on a fraction of scenes, due to its long running time,
so we exclude the curve for this method.

In~\cref{fig:fig_qual_results} we show reconstructions for several scenes.
The top part of the figure shows the results produced by methods of different types:
an RGB-only UniMVSNet,
an RGB-only NeuS with neural representation,
Neural RGB-D surface reconstruction,
and a depth-only TSDF Fusion.
The middle part shows the best result per scene \wrt.~Recall.
The bottom part shows the best result \wrt.~Precision.

In these figures we show the measures calculated with the threshold of 0.5\,mm.
In the supplementary material we show distributions of the measures for other values of the threshold,
distributions of additional quality measures, and additional reconstruction visualizations.

\noindent \textbf{Discussion.}
The best method \wrt~Recall (ACMP)
reconstructs all scenes in our dataset on at least 53\%, with the distance threshold of 0.5\,mm,
but only half of the scenes on 80\% or more.
The best method \wrt~Precision and the overall quality represented with F-score (VisMVSNet)
reconstructs all scenes with at least 32\% accuracy and only 14 scenes with accuracy higher than 80\%.
This demonstrates that our dataset contains plenty of challenges for state-of-the-art 3D reconstruction methods.
In particular, featureless parts of the surface, especially with sharp reflections,
are often missing in the reconstruction, as illustrated in~\cref{fig:fig_qual_results}.
At the same time, VisMVSNet significantly outperforms UniMVSNet \wrt~Precision and F-score and performs almost as well \wrt~Recall.
This is opposite to their relative performance on prior benchmarks (TnT and DTU),
which indicates that our dataset poses a different set of challenges.

Comparison of the methods of different types shows that reconstructions from depth maps are significantly less accurate
than reconstructions from RGB images, although sometimes may be more complete,
as illustrated at the top of~\cref{fig:fig_qual_results}.
This demonstrates that these modalities can complement each other.
Similarly, NeuS, which uses a neural surface representation,
fills-in the areas of the surface challenging for RGB-based methods, but often inaccurately.
Remarkably, Neural RGB-D surface reconstruction produces the surface of a significantly lower quality in comparison to VisMVSNet and NeuS,
while using the same input RGB images and additionally the depth maps.
This illustrates that there is a room for development of 3D reconstruction methods that effectively use both modalities,
and we believe that our dataset will facilitate the development of such methods.

\section{Conclusion}
\label{sec:conclusions}
We presented a new dataset for evaluation and training of 3D reconstruction algorithms.
Compared to prior datasets the distinguishing features of ours include
a large number of sensors of different modalities and resolutions, depth sensors in particular,
selection of scenes presenting difficulties for many existing algorithms,
and high-quality reference data for these scenes.
Our dataset can support training and evaluation of methods for many variations of 3D reconstruction tasks,
in particular, learning-based 3D surface reconstruction from multi-view RGB-D data.

The main (intentional) limitation of our dataset is the use of the laboratory setting:
the focus on static isolated objects with easy-to-separate background,
the same camera trajectory for all scenes,
the laboratory lighting.
Another possible limitation is a small range of object sizes, limited by the physical size of the setup.

\vspace{7pt plus 1pt minus 1pt}  
\paragraph{Acknowledgements.}
The authors acknowledge the use of Skoltech supercomputer Zhores~\cite{zacharov2019zhores} for obtaining the results presented in this paper.
E. Burnaev and O. Voynov were supported by the Analytical center under the RF Government
(subsidy agreement 000000D730321P5Q0002, Grant No. 70-2021-00145 02.11.2021).
\vfill
\begin{figure*}[p]
\centerline{\includegraphics[width=\textwidth]{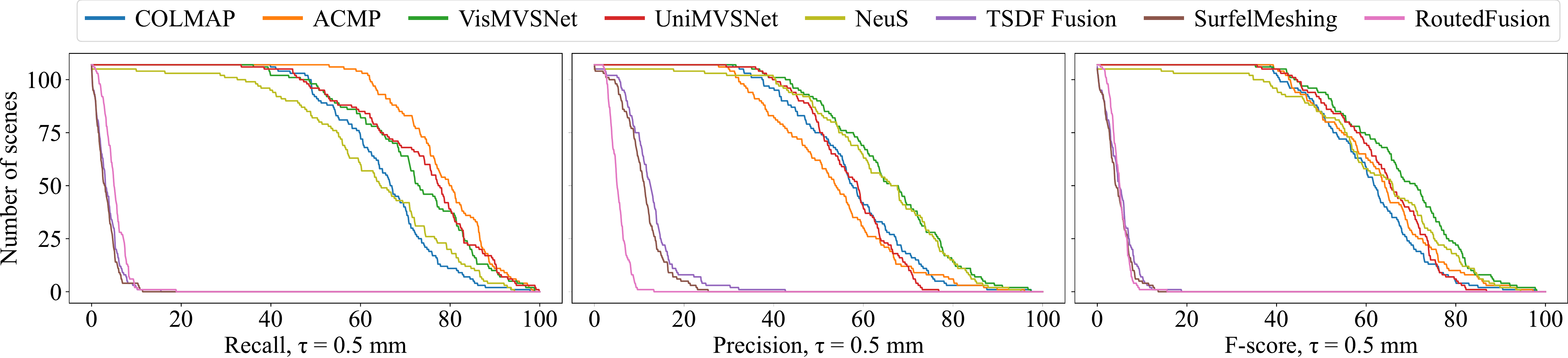}}
\caption{
\textbf{Average performance per method}
as the number of scenes with reconstruction quality better than the value on the X~axis.
}
\label{fig:fig_measures_vs_scenes_n}
\end{figure*}

\def\skipv{.4em}
\begin{figure*}[p]
\centerline{\includegraphics[width=\textwidth]{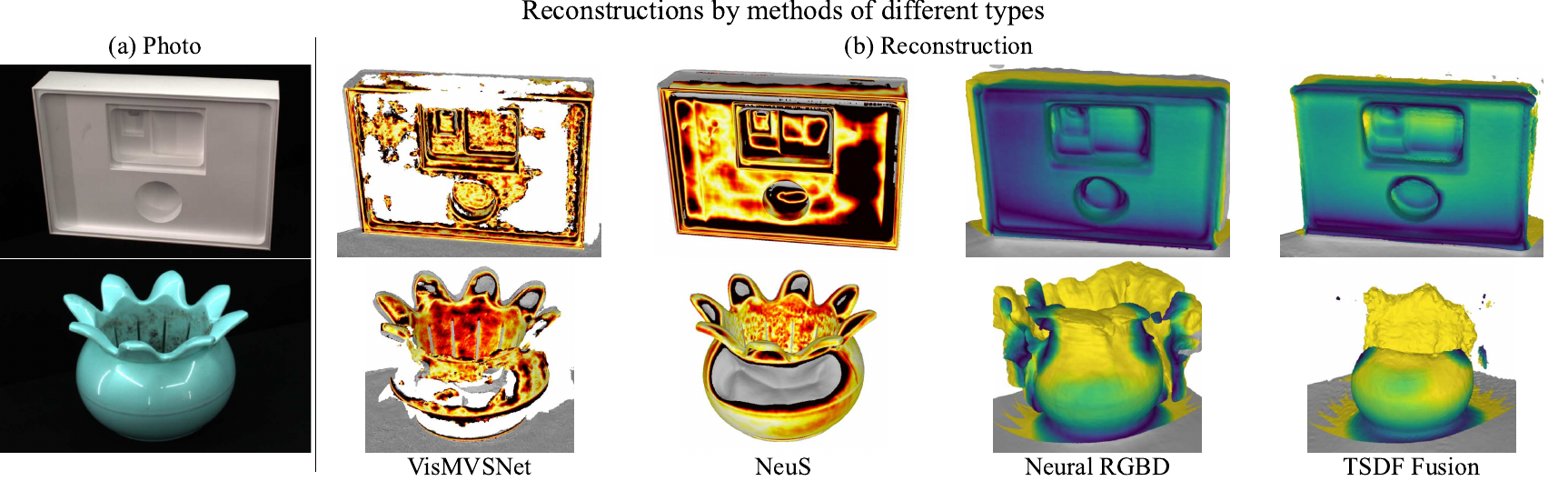}}
\vskip\skipv
\centerline{\includegraphics[width=\textwidth]{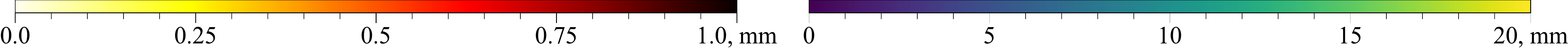}}
\vskip\skipv
\centerline{\includegraphics[width=\textwidth]{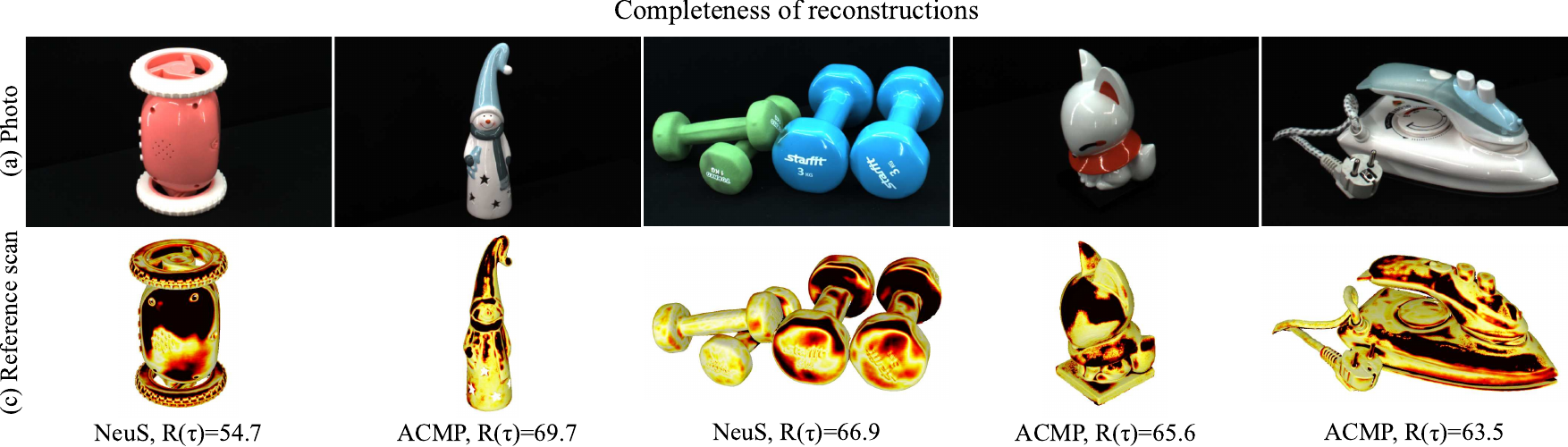}}
\vskip\skipv
\centerline{\includegraphics[width=\textwidth]{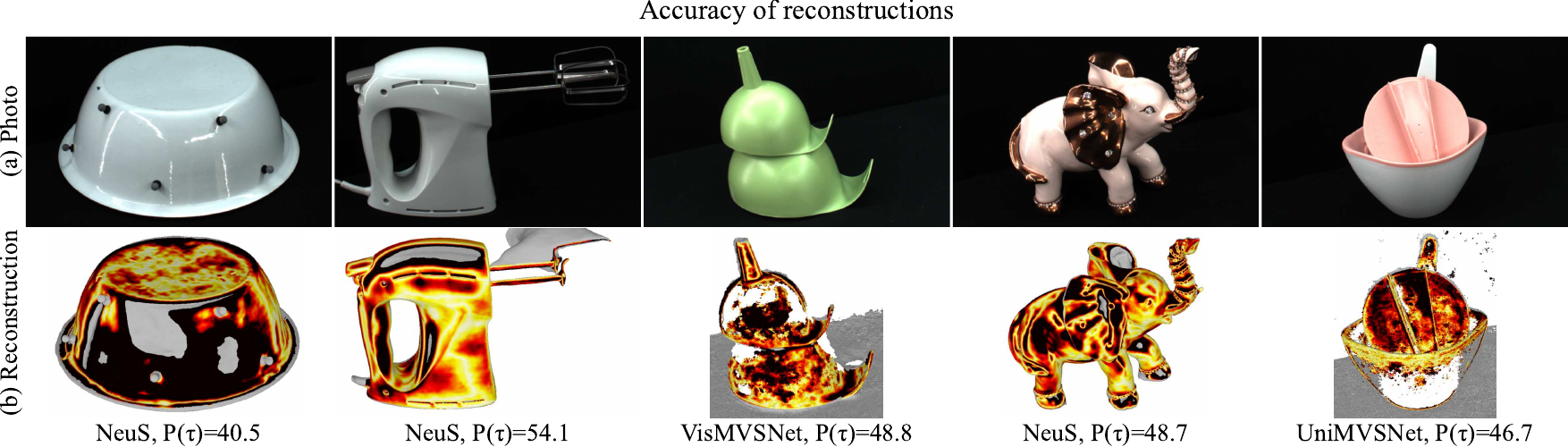}}
\caption{
\textbf{Qualitative results.}
(a) A photo of the scene.
(b) Reconstruction with color-coded distance to the SL scan.
(c) The SL scan with color-coded distance to the reconstruction.
The two bottom parts show only the best result per scene and the respective value of Recall or Precision for \(\tau=0.5\)~mm.
Grey color represents undefined distance to the real surface, as explained in the supplementary material.
}
\label{fig:fig_qual_results}
\end{figure*}

        \iftoggle{separatesupp}{
    \title{Multi-sensor large-scale dataset for multi-view 3D reconstruction\\\ Supplementary material}
    \maketitle
}{
    
\begin{figure*}[tp]
\centerline{\includegraphics[width=\textwidth]{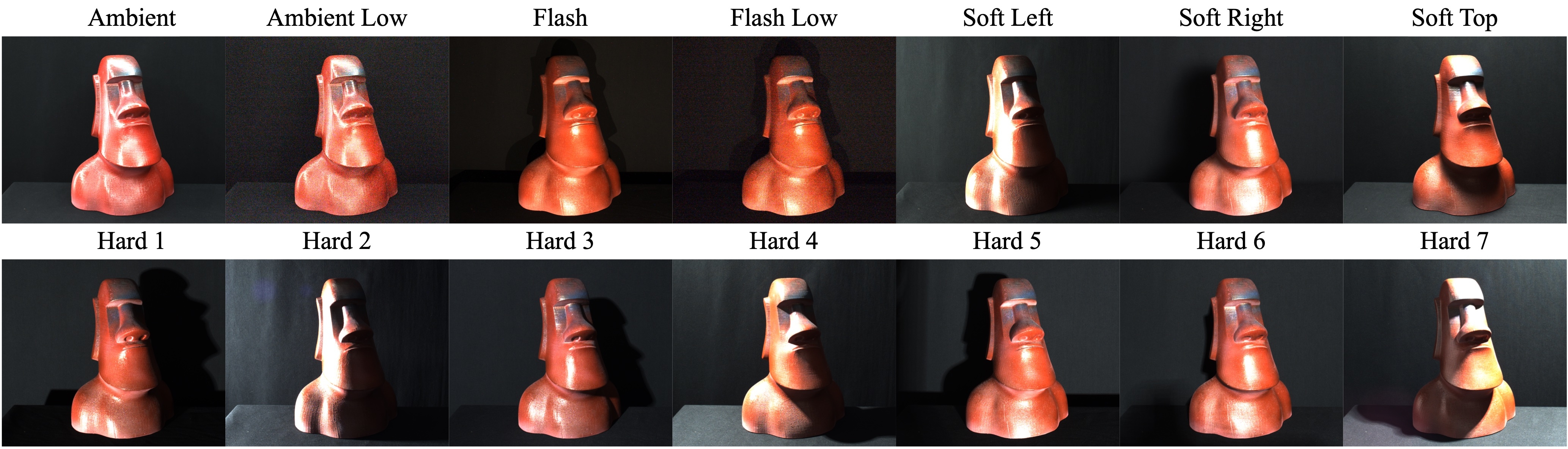}}
\caption{
\textbf{Lighting setups in our dataset:}
ambient lighting, a flashlight attached to the camera, three variants of soft light coming from different directions,
and seven variants of hard light.
For the ambient and the flash lighting there is an additional low exposure variant:
notice a higher noise level in Ambient Low and Flash Low.
}
\label{fig:fig_all_light_variations}
\end{figure*}

    \section*{Supplementary material}  
}
\appendix

In~\cref{tab:tbl_dataset_comparison_extended} we show an extended comparison of our dataset to a number of relevant datasets.
In~\cref{sup:lighting,sup:sensors} we discuss the choice of lighting variations in our dataset, and the choice of sensors.
In~\cref{sup:acquisition,sup:calibration} we provide details of data acquisition and camera calibration processes.
In~\cref{sup:train_test_methods,sup:evaluation} we provide details of testing and evaluation of 3D reconstruction methods on our dataset.
In~\cref{sup:results} we show complete evaluation results.
Finally, in~\cref{sup:material_properties} we describe and discuss the variability of key surface reflection parameters in our dataset.

\section{Lighting variability}
\label{sup:lighting}

In \cref{fig:fig_all_light_variations} we illustrate all 14 lighting setups in our dataset.
\emph{Ambient Low} and \emph{Flash Low} correspond to \emph{real-time / high-noise} camera settings
for the ambient diffuse lighting and the phone flashlight.
We aimed to provide a broad range of realistic lighting conditions:
directional light sources and flashlights of the phones provide eight samples of \textquote{hard} light,
typical, for example, for streetlight;
soft-boxes provide three samples of diffuse light, typical for indoor illumination;
LED strips imitate ambient light, typical for cloudy weather.

\section{Sensors}
\label{sup:sensors}

We aimed to include commodity RGB-D sensors with different properties.
Smartphones are ubiquitous, and are increasingly commonly augmented with a depth sensor,
while Kinect and RealSense devices represent dedicated RGB-D cameras.
We included smartphones that capture depth with a time-of-flight sensor,
Kinect~v2 also uses a time-of-flight sensor but with a higher resolution and accuracy,
and the RealSense device uses stereo-matching of infra-red images (a different technology).
These devices capture depth maps with different resolution, level of noise, and different artefacts,
as briefly illustrated in~Figure~4.
The structured-light scanner provides the reference 3D data for these devices.

These devices include commodity RGB sensors with different resolutions;
we supplemented them with industrial RGB cameras with high-quality optics and low-noise sensors.
The pair of industrial RGB cameras can serve as yet another source of depth maps based on stereo-matching of RGB images,
in contrast to IR images in RealSense.

The data captured in the same environment with different sensors can be used
to test generalization ability of computer vision methods,
or to train a generator of synthetic data to reproduce a specific sensor, \etc.

\iftoggle{separatesupp}{
\begin{figure*}[tp]
\centerline{\includegraphics[width=\textwidth]{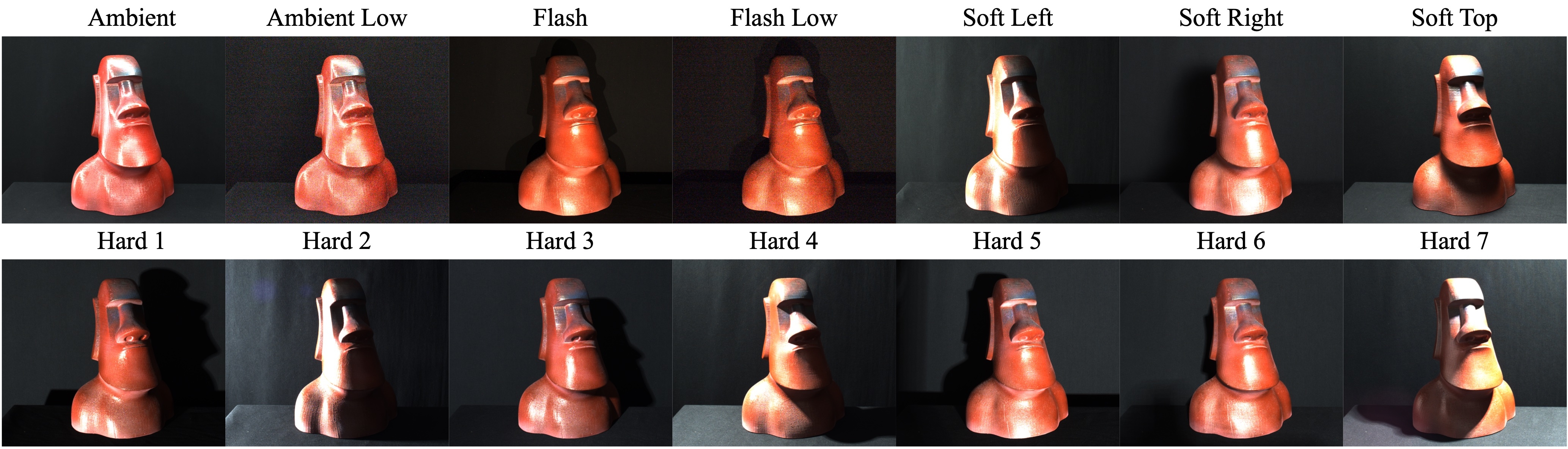}}
\caption{
\textbf{Lighting setups in our dataset:}
ambient lighting, a flashlight attached to the camera, three variants of soft light coming from different directions,
and seven variants of hard light.
For the ambient and the flash lighting there is an additional low exposure variant:
notice a higher noise level in Ambient Low and Flash Low.
}
\label{fig:fig_all_light_variations}
\end{figure*}
}{}
\nottoggle{separatesupp}{\begin{figure*}[tp]
\centerline{\includegraphics[width=\textwidth]{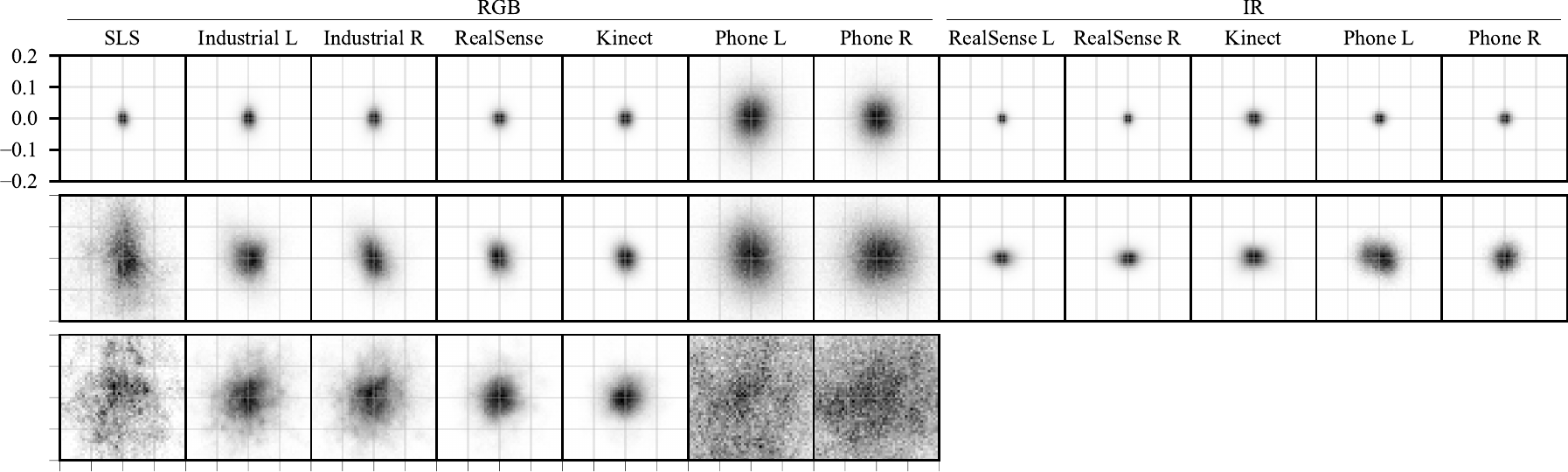}}
\vskip0.4em
\centerline{\includegraphics[width=\textwidth]{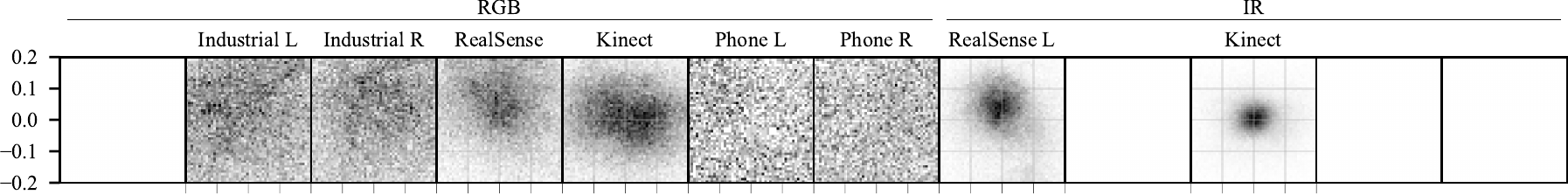}}
\caption{
\textbf{Error distributions for camera calibration,}
as histograms of 2D differences between the detection of a feature on the calibration pattern
and its projection from 3D to the image.
The range of each histogram is \([-0.2, 0.2]\) pixels in each dimension.
Each histogram represents one sensor at one of the three steps of the calibration procedure:
estimation of intrinsic camera models (first row),
estimation of relative position of the sensor within the rig (second row),
estimation of position of the sensor on the scanning trajectory (third row).
The histograms for the IR sensors in the third row are not shown,
since the trajectory of the IR sensors is calculated directly from the trajectory of their RGB companions.
The fourth row shows the errors for the baseline approach of estimation of relative positions of the sensors within the rig
(compare with the second row).
}
\label{fig:fig_cam_calib_hists}
\end{figure*}
}{}
\section{Data acquisition details}
\label{sup:acquisition}
\paragraph{Lens settings.}
We set the focal distance for the industrial RGB cameras and for the cameras of the phones
to the expected average distance from the cameras to the surface of the scanned object, namely 62.5 cm.
To extend the depth of field for the industrial cameras to the whole scanning area,
we set the aperture to the minimal value that does not cause any visual blur due to diffraction.
We kept the lens parameters for Kinect and RealSense at their factory settings.

\paragraph{White balance.}
We set the white balance for all RGB sensors fixed at the start of each scanning session (\ie., daily),
using a black-and-white calibration pattern.
Most of our light sources have the same light temperature so that their light appears white under this setting,
except for the ambient illumination which has a somewhat higher green component,
and for the flashlight of the phones which has a yellow tint.
We did not fix the white balance setting for Kinect, for which this control is not available.

\paragraph{Exposure and gain.}
For Kinect, it is not possible to control camera exposure and gain directly.
Instead, the built-in auto-exposure function is permanently turned on,
which sets the exposure and gain so that the mean pixel value over the whole image is 50\% gray.
Since in our setup a large portion of the image is the black background,
the built-in algorithm produces over-exposed images.
To minimize this effect, we added a dimming light filter mounted on a servomotor to the rig,
which is placed in front of the RGB camera of Kinect automatically when the bright light sources are activated,
and removed for the dim ones.

\section{Camera calibration details}
\label{sup:calibration}
The original implementation of the calibration pipeline of~\cite{schops2020having} supports calibration of rigid camera rigs,
however, a straightforward application of this pipeline for our camera rig in its entirety
proved to be numerically unstable due to the properties of the setup.
Firstly, the included sensors have a large variation in
the field of view (from 30\textdegree{} for the SLS camera to 90\textdegree{} for the IR sensors of RealSense)
and resolution (from 0.04\,MPix for the IR sensors of the phones to 40\,MPix for their RGB sensors).
Secondly, the focus of the cameras of the phones, being fixed programmatically, fluctuates slightly over time,
which we relate to thermal deformations of the device
(see~\cite{elias2020assessing} for a study of such an effect).
Finally, the camera rig deforms slightly depending on its tilt in different scanning positions.
To avoid the loss of accuracy we split the calibration procedure into several steps.

First, we obtained intrinsic camera models for each sensor independently.
Then, for the sensors with a relatively high resolution and stable focus,
namely the SLS and all RGB sensors except the sensors of the phones,
we estimated their relative position within the rig for its vertical orientation.
Next, we estimated the relative position of RGB sensors of the phones within the rig, \wrt~the other sensors.
After that, we estimated the position of each IR (depth) sensor \wrt~the RGB sensor of the respective device,
assuming that the sensors are fixed rigidly to the frame of the device.
Finally, we estimated the position of each RGB sensor individually for different scanning positions of the robotic arm,
and then linked the positions of all sensors together through a scanning position with the vertical orientation of the rig.

This whole procedure required capturing thousands of images of the calibration pattern,
which we almost fully automated with the use of the robotic arm,
except for several manual reorientations of the pattern.

In~\cref{fig:fig_cam_calib_hists} we show distributions of calibration error for each sensor at different stages
of the calibration procedure.
Note that the resolution of the RGB sensors of the phones is relatively high compared to the other RGB sensors
so a higher value of the calibration error measured in pixels is expected.

At the bottom of~\cref{fig:fig_cam_calib_hists} we show calibration errors for the straightforward application
of the calibration pipeline to our camera rig.
Here, we estimated the relative positions for a subset of sensors within the rig simultaneously.
The mean error of the straightforward approach is 1.8-5.7 times higher depending on the sensor compared to ours.

\iftoggle{separatesupp}{\begin{figure*}[tp]
\centerline{\includegraphics[width=\textwidth]{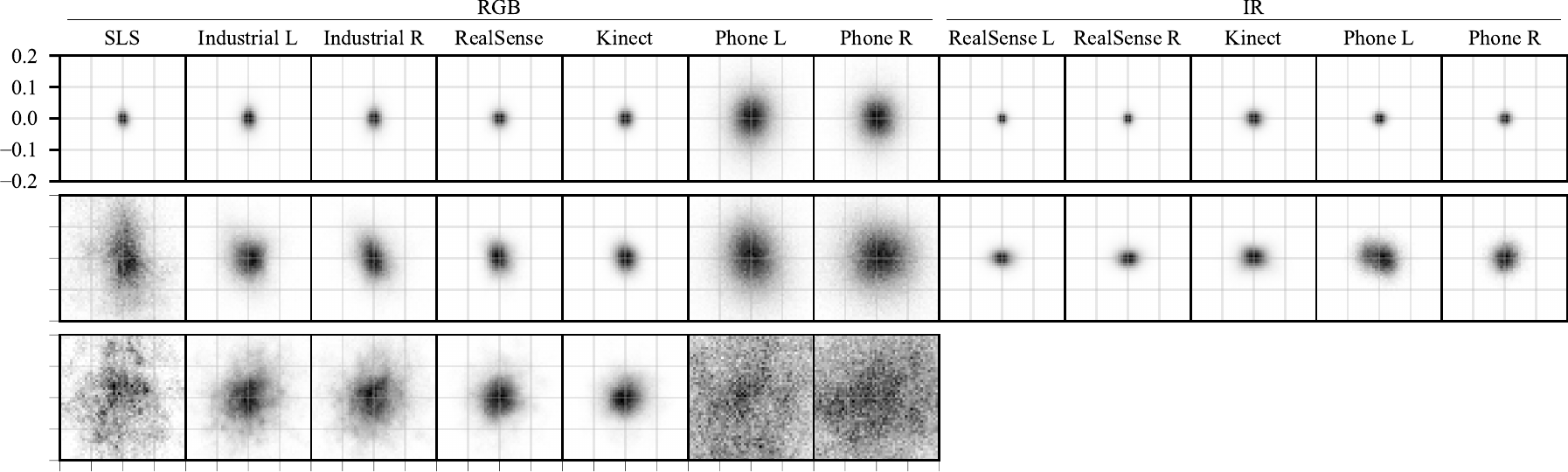}}
\vskip0.4em
\centerline{\includegraphics[width=\textwidth]{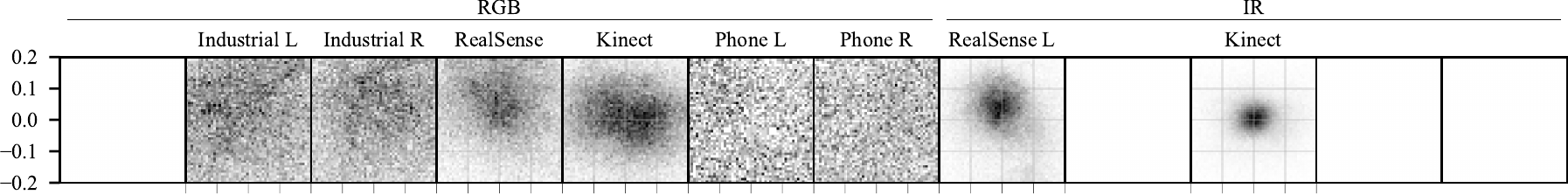}}
\caption{
\textbf{Error distributions for camera calibration,}
as histograms of 2D differences between the detection of a feature on the calibration pattern
and its projection from 3D to the image.
The range of each histogram is \([-0.2, 0.2]\) pixels in each dimension.
Each histogram represents one sensor at one of the three steps of the calibration procedure:
estimation of intrinsic camera models (first row),
estimation of relative position of the sensor within the rig (second row),
estimation of position of the sensor on the scanning trajectory (third row).
The histograms for the IR sensors in the third row are not shown,
since the trajectory of the IR sensors is calculated directly from the trajectory of their RGB companions.
The fourth row shows the errors for the baseline approach of estimation of relative positions of the sensors within the rig
(compare with the second row).
}
\label{fig:fig_cam_calib_hists}
\end{figure*}
}{}
\section{Parameters of tested methods}
\label{sup:train_test_methods}

We tested all methods using ground-truth camera poses and intrinsic camera models,
after the refinements of camera poses and depth camera calibration.
We used RGB images from the right industrial camera and the depth maps from Kinect
at full resolution after removing distortion (which resulted in a small amount of cropping at the boundaries),
specifically, \(2368\times 1952\) for RGB and \(496\times 400\) for depth.

For each method, we tried to pick the values of the method parameters which resulted in the best reconstructions on our dataset,
based on 5-10 typical scenes.
We describe the parameters using the original notations from the respective works.

We tested \textbf{COLMAP}~\cite{schoenberger2016sfm-colmap,schoenberger2016mvs-colmap,colmap_url}
with parameters set to their \textquote{performance} values recommended in the software documentation.
Specifically, for feature extraction, we enabled estimation of affine shape of SIFT features,
and enabled the more discriminative DSP-SIFT features instead of plain SIFT;
for feature matching, we disabled estimation of multiple geometric models per image pair;
for patch-match stereo, we enabled the regularized geometric consistency term;
finally, for stereo fusion we set the minimum number of fused pixels to produce a point to 3,
the maximum relative difference between measured and projected pixels to 1,
and the maximum depth error to 1\,mm.

We tested \textbf{ACMP}~\cite{xu2020ACMP,acmp_url}
with the default parameters used in the source code.
For this MVS method and the two methods below
we sampled the depth hypotheses uniformly from 473\,mm to 983\,mm with a 2\,mm resolution,
and for view selection used the strategy proposed in~\cite{yao2018mvsnet},
applied to the reference SL scan instead of a sparse reconstruction of the scene.

To test \textbf{VisMVSNet}~\cite{zhang2020vismvsnet,vismvsnet_url}
we trained it from scratch on BlendedMVG dataset~\cite{blendedmvg_url},
which is an extended version of the BlendedMVS dataset~\cite{BlendedMVS:yao2020blendedmvs},
originally used by the authors of the method.
We used the original training parameters and trained the network on a single Nvidia GTX 1080Ti GPU
for 2 epochs with a batch size of 2 (342K iterations),
using Adam optimizer~\cite{kingma2015adam} with a learning rate of \(10^{-4}\).

We tested VisMVSNet with
the number of neighboring source images \(N_v = 7\),
the numbers of depth hypotheses \(N_{d,1}, N_{d,2}, N_{d,3} = 64, 32, 16\),
the minimal number of consistent views during fusion \(N_f=4\),
and the fusion probability thresholds \(p_{t,1}, p_{t,2}, p_{t,3} = 0.8, 0.7, 0.8\).

We tested \textbf{UniMVSNet}~\cite{peng2022rethinking,unimvsnet_url}
using the  published model trained on DTU dataset~\cite{DTU:jensen2014large} and fine-tuned on BlendedMVS dataset.
We used the number of neighboring source images \(N = 11\),
the numbers of depth hypotheses \(M_1, M_2, M_3 = 64, 32, 16\),
and the fusion probability thresholds \(\phi_1, \phi_2, \phi_3 = 0.1, 0.15, 0.9\).

We tested \textbf{NeuS}~\cite{wang2021neus,neus_url}
with the original hyperparameters, optimizing the network for 300K iterations.
To accelerate convergence and prevent oversmoothing of reconstruction
we cropped the input images for this method to the bounding box of the object expanded by \(\sim 10\)~pixels.

To test \textbf{TSDF Fusion}~\cite{curless_volumetric_1996,izadi_kinectfusion:2011}
we used an implementation of this algorithm from Open3D library~\cite{zhou2018open3d},
with a voxel size of 3\,mm, and a TSDF truncation distance of 2\,cm.
We additionally tested an implementation of this algorithm from~\cite{niessner2013real,voxelhashing_url}
and obtained very similar results; we do not report them.

We tested \textbf{SurfelMeshing}~\cite{schops_surfelmeshing:2020,surfelmeshing_url}
with the number of inliers for depth filtering set to 1,
the depth map erosion radius set to 0.1,
and with 1 iteration of median filtering.

To test \textbf{RoutedFusion}~\cite{weder_routedfusion:2020,routedfusion_url}
we trained it from scratch on ModelNet dataset~\cite{ModelNet:CVPR15_Wu}.
We used the original training parameters for ShapeNet dataset~\cite{shapenet2015},
with an increased level of synthetic noise \(\sigma=0.01\), as suggested by the authors.
We tested RoutedFusion with a grid resolution of \(384^3\), which corresponds to a voxel size of 2.6\,mm.

We tested \textbf{Neural RGB-D Reconstruction}~\cite{azinovic2022neural,neuralrgbd_url}
with the original hyperparameters,
optimizing the network until convergence for 200K iterations,
and sampling the coarse points on the ray for every 2\,mm.

\section{Evaluation details}
\label{sup:evaluation}

We evaluated 3D reconstruction methods using
\emph{precision} \(P\parens{\tau}\) defined as the percentage of reconstruction points
which are closer to the reference surface than a distance threshold \(\tau\);
\emph{recall} \(R\parens{\tau}\) defined as the percentage of reference points
which are closer to the reconstruction than the distance threshold;
and \emph{F-score} \(F\parens{\tau}\) defined as the harmonic mean of precision and recall.
Additionally, we calculated the mean distance from the reconstruction to the reference,
and the mean distance from reference to reconstruction,
which we report in~\cref{sup:results}.

To calculate the measures based on the distance from the reference to the reconstruction, we used vertices of the full SL scan;
their average distance from the nearest neighbor is around 0.15~mm.
To evaluate the methods which reconstruct the surface in the form of a triangular mesh,
we sampled points from the mesh uniformly at a sampling distance of 0.1~mm.

The reference SL scans are incomplete, so the distance from some reconstruction points to the SL scan
does not represent the distance to the real surface of the object,
specifically, if the point lies near the missing part of the surface.
To calculate the measures using only the points for which the distance to the real surface is reliable
we used the approach of~\cite{ETH3D:schops2017multi}, and extended it using our new occluding surface.

First, we only kept the points which lie in the free space between the SL scanner and the object or near the surface of the object: we checked if the depth of the point \wrt~SL scanner for any of its scanning positions is less than the depth given by the SL scan,
plus a small tolerance
\(t_\mathrm{subsurf}=3\)~mm
to keep the points just below the surface for evaluation.

Next, to evaluate the precision metric, we checked if a point is closer to the real surface of the object than a distance threshold.
For every reconstructed point, we calculated the distance to the SL scan and the distance to the occluding surface.
If the distance to the SL scan was below the threshold, we considered the point to be closer to the real surface than the threshold.
If the distance to the occluding surface was above the threshold, we considered the point to be farther from the real surface than the threshold.
In any other case
(in which the point can only be closer to the occluding surface and farther from the SL scan than the threshold,
since the occluding surface encloses the SL scan by definition),
we considered the distance from the point to the real surface to be unknown and excluded the point from the calculation of the measure.

To visualize the distance from the reconstruction to the reference and to calculate the mean distance we used a similar strategy
and considered the distance to the real surface to be unknown
whenever the point was closer to the occluding surface than to the SL scan.
To account for the approximate nature of our occluding surface and to prevent computational instabilities at points
where it must coincide with the SL scan exactly in case of exact calculations,
we replaced the distance to the occluding surface in all calculations by its value increased by
\(\varepsilon_\mathrm{occ}=0.1\)~mm.

\section{Complete evaluation results}
\label{sup:results}

\iftoggle{separatesupp}{
    In~Figures~\figcolor{5}~to~\figcolor{111}\footnote{
        See at \href{http://skoltech3d.appliedai.tech/data/skoltech3d_supp_results.pdf}{skoltech3d.appliedai.tech/data/skoltech3d\_supp\_results.pdf}}
}{
    In~Figures in a separate PDF file\footnote{
        See at \href{http://skoltech3d.appliedai.tech/data/skoltech3d_supp_results.pdf}{skoltech3d.appliedai.tech/data/skoltech3d\_supp\_results.pdf}}
}
we show qualitative evaluation of 3D reconstruction methods.
For Neural RGB-D surface reconstruction we show the results only for four scenes:
two relatively easy ones, based on performance of all the methods,
\texttt{dragon} and \texttt{\detokenize{small_wooden_chessboard}},
and two relatively hard ones \texttt{\detokenize{green_flower_pot}} and \texttt{\detokenize{white_box}}.

In each figure in the first column titled \emph{Reconstruction}, we show the reconstruction produced by each method,
and at the bottom of this column we show the reference surface from the SL scanner.
In the column \emph{Accuracy on reconstruction}, we show the reconstructed surface  with color-coded distance to the reference surface;
at the bottom of this column we show a photo of the scene.
In the column \emph{Accuracy on reference},
we show the reference surface, with the color showing the distance from the reconstruction to the reference.
For each vertex of the reference surface, the distance is averaged over the reconstructed points for which this vertex is the closest one.
Thanks to this projection of the distance values from the reconstructed points to the reference surface,
these images show the accuracy of all points not just the ones closest to the camera.
In the column \emph{Completeness on reference},
we show the reference surface with color-coded distance to the reconstructed surface.

We use two colormaps for two different scales of error.
In the last three columns,   the points with no definite value of the distance are grey.

Since the methods TSDF Fusion, RoutedFusion, SurfelMeshing, and Neural RGB-D surface reconstruction
produce the surface with a significant error, and in particular deep below the reference surface,
we increased the value of sub-surface tolerance \(t_\mathrm{subsurf}\) for these methods from 3~mm to 20~mm.

\iftoggle{separatesupp}{
    In~Figures~\figcolor{112}~to~\figcolor{218}\textsuperscript{\textcolor{red}{1}}
}{
    In~Figures in a separate PDF file\textsuperscript{\textcolor{red}{1}}
}
we show the recall, precision, and F-score curves for reconstructed surfaces produced by the methods for each scene.
Each curve represents the measure in percent for different values of the distance threshold \(\tau\) in millimeters.
Additionally, we mark the mean distance from the reference to the reconstruction for each method
with a vertical dashed line on the recall plot,
and the mean distance from the reconstruction to the reference with a vertical dashed line on the precision plot.
Note that for some methods these lines may be out of the plot range.

Finally, in~\cref{fig:fig_comp_bars,fig:fig_acc_bars} we show the values of the recall and precision metrics
calculated with the distance threshold \(\tau=0.5\)~mm
for the reconstructions produced by the RGB-based methods COLMAP, ACMP, VisMVSNet, UniMVSNet, and NeuS.
The value of the measure for each method is represented by the right edge of the bar with the respective color.
The scenes in each figure are sorted by the best result on the scene.

\section{Material properties in our dataset}
\label{sup:material_properties}

We provide more information on the qualitative descriptors of surface reflection parameters assigned to each object and their relation to performance indicators for various reconstruction methods.

\noindent \textbf{Identification of surface properties.}
As outlined in the main text, all labels were assigned to objects by visually inspecting photos of each object, to provide a preliminary assessment of the importance of various factors for reconstruction quality.
Multiple photos of each object were used, with backlight proving particularly useful to establish the degree of translucency. The labels refer to the \emph{dominant} material or materials of the objects, which have most influence on integral metrics of reconstruction quality.
While this classification is imprecise and not a substitute for photometric measurements, as we use only rough estimation for each property, typically 3 buckets, we expect that visually assigned labels are highly correlated with the actual physical parameters.
Where necessary, more than one label was assigned (\eg, for objects consisting of multiple parts with distinct reflectivity properties, with no single dominant material); in this case several   labels for some properties are used simultaneously.

Most of our labels are aligned with typical parameters of simple reflectance models: diffuse and specular reflection coefficient, shininess/roughness, measuring the reflection peak width, and translucency. 

\begin{itemize}
\item \emph{Translucency. } We assess the degree of translucency for the  materials of the large parts of the object, from \emph{none} (the default), through \emph{low, medium, high,} all the way to completely \emph{transparent.}

\item \emph{Reflection sharpness. } We characterize how sharp the reflectance function peak is,
if highlights are present on the object (i.e., it is not purely diffuse), grouping the objects into four categories, from low to very high. 

\item \emph{Specularity. } We visually estimate the ratio of specular to diffuse reflection for the dominant object materials, resulting in a  degree of manifestation of view-dependent highlights, complementing \emph{mirror-like} tag with a weaker label.
We distinguish between no view dependent highlights \emph{(diffuse),} largely diffuse highlights \emph{(low),} somewhat diffuse light reflections and wide highlights  \emph{(medium),} and narrow highlights for partially mirror-like surfaces where one can only see sharp reflections of lights \emph{(high).}

\item \emph{Mirror-like. } A binary tag for surfaces for which, in addition to a near-perfect reflection peak, the diffusive component is low relative to specular. This label overlaps very high reflection sharpness labels.

\item \emph{Metallic. } A binary tag for surfaces made of metal; compared to dielectric surfaces, these surfaces are always completely non-transparent, and the specular component of reflected light tend to have the same color as diffuse, while for dielectric materials it is closer to the color of the incident light. 

\item \emph{Geometric features. }  We visually assess the dominant qualitative scale of geometric structures of the surface, if any. 
We seek to distinguish between fine 3D structure with characteristic scale close but above SLS 3D resolution \emph{(small),} a larger, discernable geometry with feature size equal to a small fraction of the object size \emph{(medium)}.  Additionally, among surfaces lacking a dominant feature scale,  we identified those with flat areas of sufficiently large size to make an impact on overall reconstruction quality (\emph{flat} label), in the absence of 3D texture.

\item \emph{Texture type. } We differentiate between several types of textured surfaces: most of the surface covered by high-frequency image texture \emph{(color),} sufficiently small scale (below or close to what the scanner can resolve), high-frequency displacement variation \emph{(3D),} or other texture-like imperfections such as dirt, speckles, or visible roughness \emph{(imperfections).}
\end{itemize}

\noindent \textbf{Correlating properties and reconstruction performance.}
Most surface reflectance features mentioned above are expected to influence performance of common reconstruction methods.

We use data-driven approach to model the variability in a given measure of reconstruction performance caused bydifferent reflectance properties.  Viewing all measures described in~\cref{sup:evaluation} as target variables, we constructed a numerical feature representation for each scene in our dataset; and  used sparse $L_1$ regularised (Lasso) regression~\cite{tibshirani1996regression} to determine how each surface reflectance feature modulates reconstruction performance in each scene.

As we wanted to determine which features had statistically the most influence on reconstruction performance, we used the automatic feature selection mechanism built into the Lasso algorithm. 

Our target variables were \emph{precision} \(P\parens{0.5\text{\,mm}}\), \emph{recall} \(R\parens{0.5\text{\,mm}}\), and \emph{F-score} \(F\parens{0.5\text{\,mm}}\) for the image-based methods ACMP, COLMAP, VisMVSNet, and UniMVSNet (\cref{sup:train_test_methods}).
For constructing the feature representation of each scene, we formed 25~binary features from the 7~properties mentioned above by encoding the presence of each surface property as 1, and its absense as 0.
For fitting the model, we used the implementation of Lasso available in~\texttt{scikit-learn}~\cite{pedregosa2011scikit}; we z-scored all features, and, for each target variable, sought to find an optimal value of the regularisation parameter $\alpha$ (a minimizer of RMSE measure) using leave-one-out cross-validation by varying it over a range $[0.01, 2.0]$, and record the values of the $R^2$ statistic, the regularization weight~$\alpha$ as well as the regression weights of the final fit. 

\cref{tab:feature_importance_scores} displays (normalized) values of coefficients weighting the (normalized) value of each feature in each constructed linear model. 
Most regression problems have demonstrated a consistent value of the regularization weight~$\alpha \in [0.18, 0.57]$ with an average of 0.36; the values of coefficient of determination $R^2$ vary from 0.24 to 0.54 with a mean of 0.43, indicating that a reasonable performance has been achieved.

\noindent \textbf{Factors affecting algorithms performance.}
Using the process described above, we have identified five labels having the most pronounced \emph{negative effect} on reconstruction quality for image-based MVS methods, according to a mean F-score across methods:
very high, high, or medium reflection sharpness, high specularity, medium-scale geometric features; additionally, a negative effect from high translucency, low specularity is also expected, unlike moderate but notable contribution of non-metallic and diffuse materials.
Conversely, having any type of texture (either color, 3d, or texture-like surface imperfections) \emph{contributes positively} to reconstruction performance as one expect for MVS-type algorithms; the same can be said about non-translucent objects. 

 We consider this study very preliminary, given the approximate nature of our labels.  Its results are encouraging as these show clear correlation between surface reflectance properties and algorithm performance.

\begin{table*}[p]
\centering
\includegraphics[width=\textwidth]{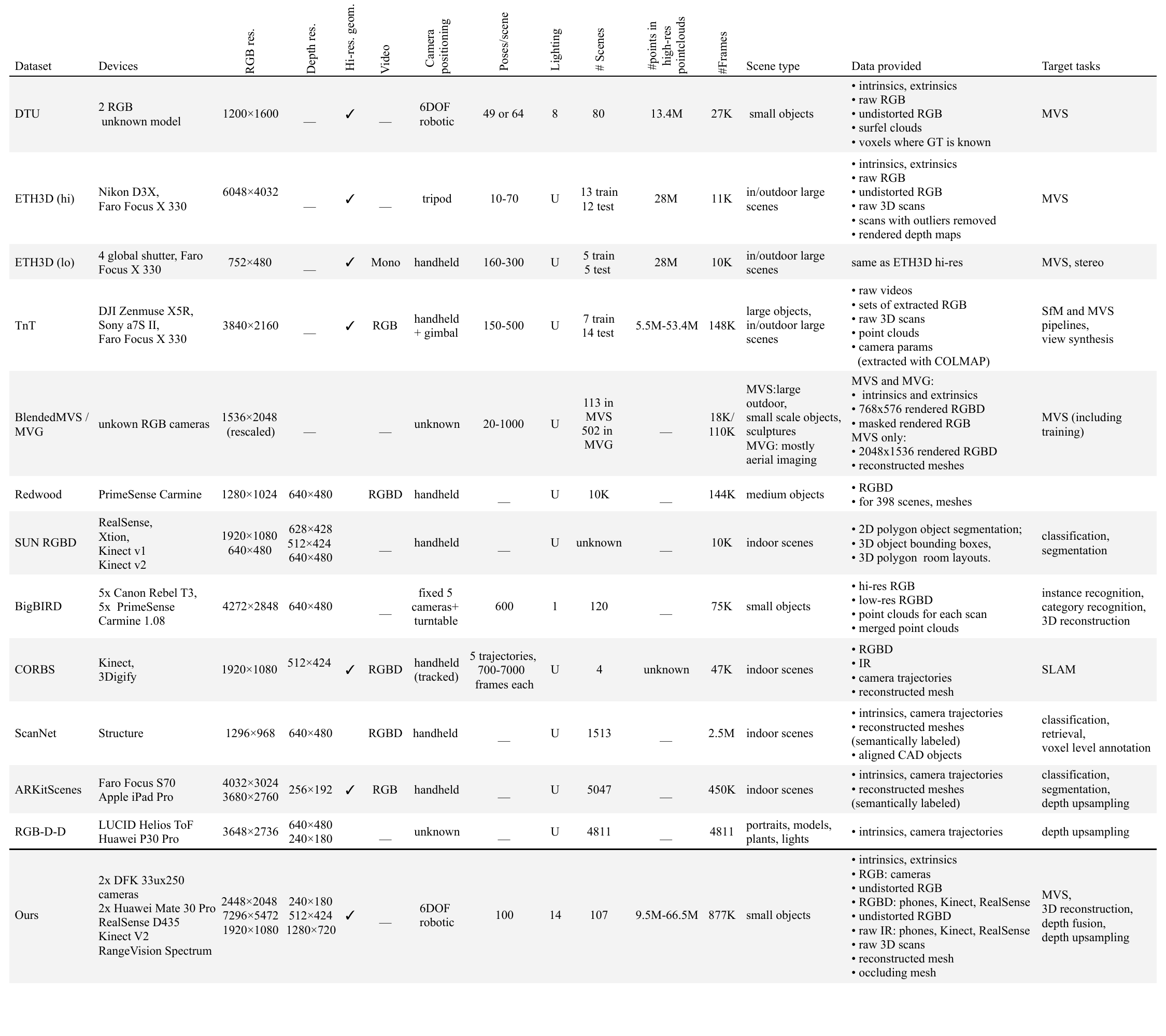}
\caption{
\textbf{Comparison of our dataset to the most widely used related datasets.}
U indicates uncontrolled lighting;
frames are counted per sensor, \ie., all data from an RGB-D sensor are counted as a single frame.
The number of separate images acquired may be considerably larger (1.4\,M for our dataset).
All scenes, from both training and testing sets, were counted.
}
\label{tab:tbl_dataset_comparison_extended}
\end{table*}

\begin{figure*}[p]
    \centerline{\includegraphics[width=\textwidth,height=0.97\textheight,keepaspectratio]{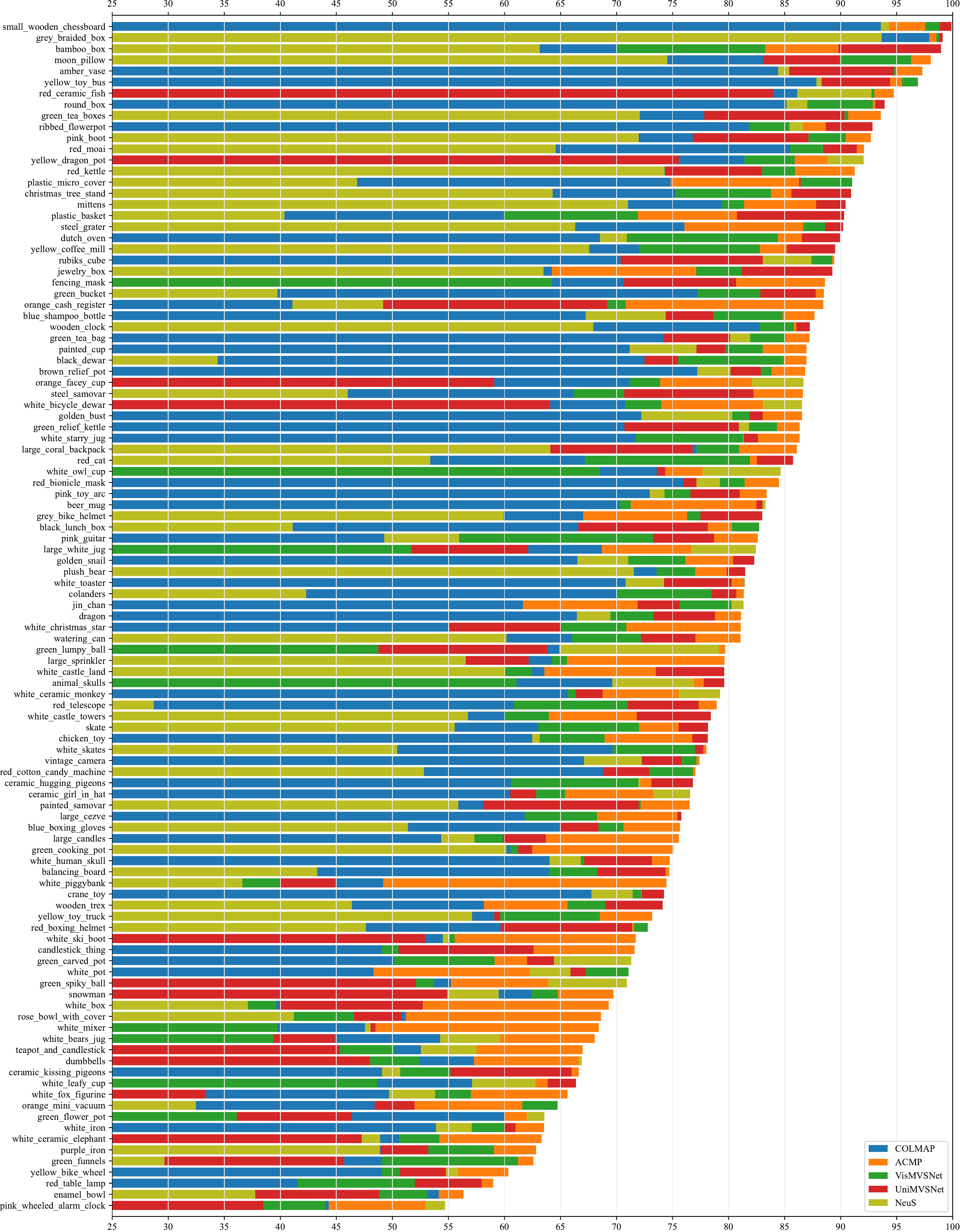}}
    \caption{\textbf{Recall for the RGB-based methods for all scenes} with the distance threshold \(\tau=0.5\)~mm.}
    \label{fig:fig_comp_bars}
\end{figure*}

\begin{figure*}[p]
    \centerline{\includegraphics[width=\textwidth,height=0.97\textheight,keepaspectratio]{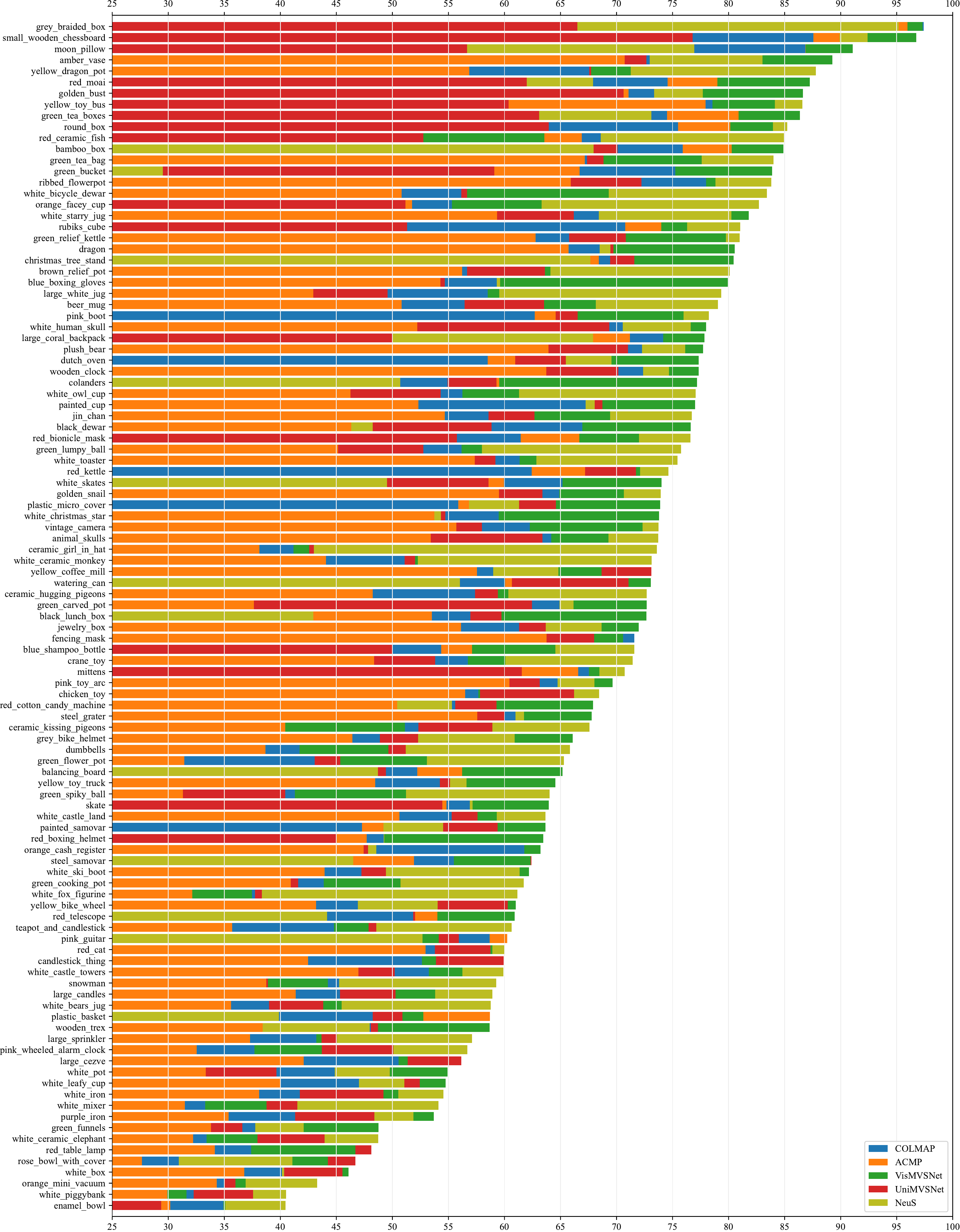}}
    \caption{\textbf{Precision for the RGB-based methods for all scenes} with the distance threshold \(\tau=0.5\)~mm.}
    \label{fig:fig_acc_bars}
\end{figure*}

\begin{table*}[p]
\centering
\includegraphics[
width=\textwidth]{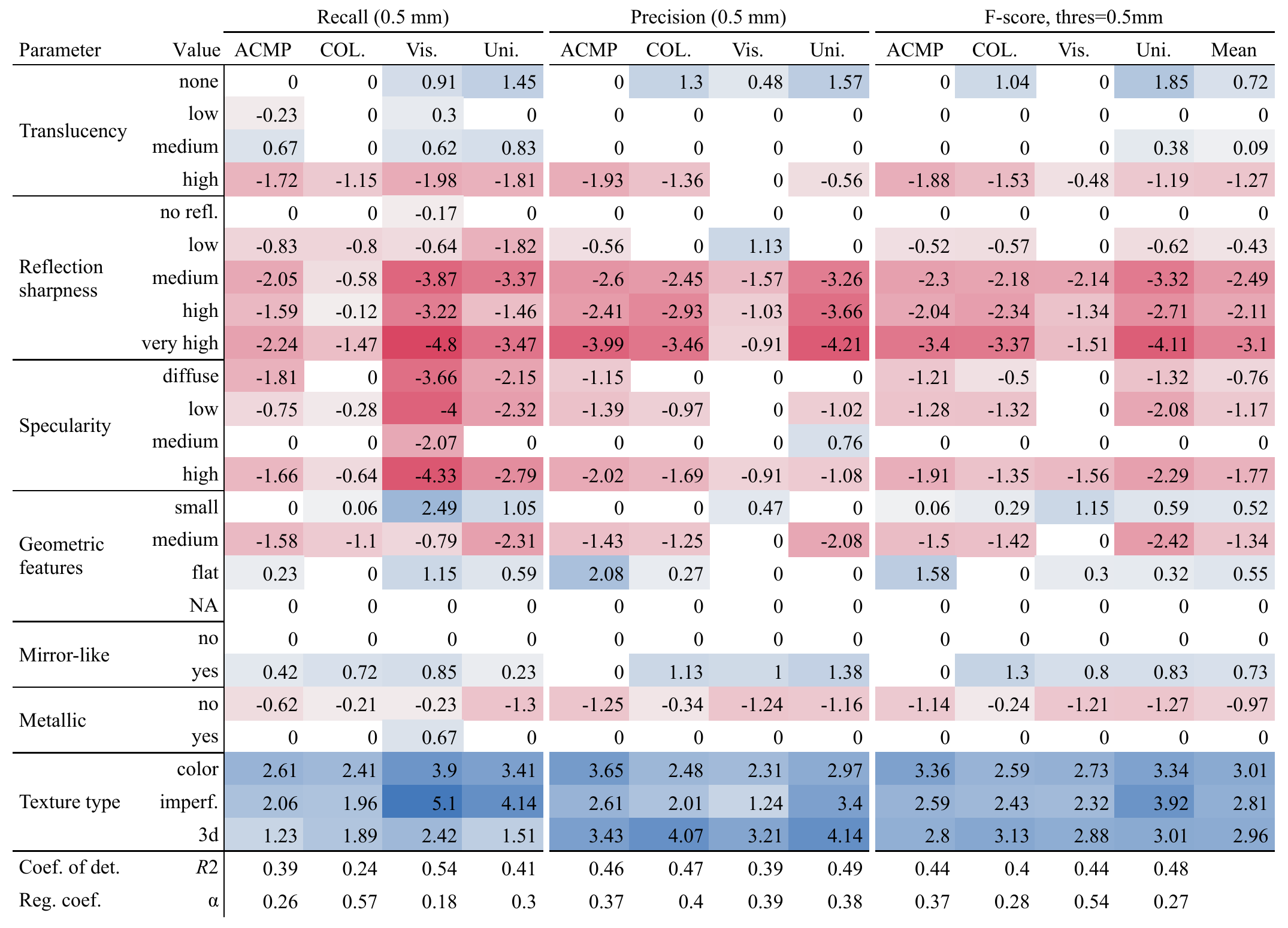}
\caption{
\textbf{Data-driven estimates of surface reflection features in our datasets.}
Note the \emph{negative effect} on reconstruction quality (rightmost column, negative numbers highlighted in \colorbox[rgb]{0.88,0.52,0.58}{red}) of very high, high, or medium reflection sharpness, high specularity, medium-scale geometric features. 
Conversely, note the \emph{positive contribution} to reconstruction performance of color textures, 3d textures, or texture-like surface imperfections (rightmost column, large positive values highlighted in \colorbox[rgb]{0.43,0.58,0.78}{blue}).
\emph{COL.}, \emph{Vis.}, and \emph{Uni.} denote COLMAP, VisMVSNet, and UniMVSNet respectively.
}
\label{tab:feature_importance_scores}
\end{table*}

        {\clearpage\newpage\small\bibliographystyle{ieee_fullname}\bibliography{src/bib}}

\begin{thebibliography}{100}\itemsep=-1pt

\bibitem{aanaes2016large}
Henrik Aan{\ae}s, Rasmus~Ramsb{\o}l Jensen, George Vogiatzis, Engin Tola, and
  Anders~Bjorholm Dahl.
\newblock Large-scale data for multiple-view stereopsis.
\newblock {\em International Journal of Computer Vision}, 120(2):153--168,
  2016.

\bibitem{aliev2020neural}
Kara-Ali Aliev, Artem Sevastopolsky, Maria Kolos, Dmitry Ulyanov, and Victor
  Lempitsky.
\newblock Neural point-based graphics.
\newblock In {\em European Conference on Computer Vision}, pages 696--712.
  Springer, 2020.

\bibitem{azinovic2022neural}
Dejan Azinovi{\'c}, Ricardo Martin-Brualla, Dan~B Goldman, Matthias
  Nie{\ss}ner, and Justus Thies.
\newblock Neural rgb-d surface reconstruction.
\newblock In {\em Proceedings of the IEEE/CVF Conference on Computer Vision and
  Pattern Recognition}, pages 6290--6301, 2022.

\bibitem{Barnes2009PatchMatch}
Connelly Barnes, Eli Shechtman, Adam Finkelstein, and Dan~B Goldman.
\newblock {PatchMatch}: A randomized correspondence algorithm for structural
  image editing.
\newblock {\em ACM Transactions on Graphics (Proc. SIGGRAPH)}, 28(3), Aug.
  2009.

\bibitem{dehghan2021arkitscenes}
Gilad Baruch, Zhuoyuan Chen, Afshin Dehghan, Tal Dimry, Yuri Feigin, Peter Fu,
  Thomas Gebauer, Brandon Joffe, Daniel Kurz, Arik Schwartz, and Elad Shulman.
\newblock {ARK}itscenes - a diverse real-world dataset for 3d indoor scene
  understanding using mobile {RGB}-d data.
\newblock In {\em Thirty-fifth Conference on Neural Information Processing
  Systems Datasets and Benchmarks Track (Round 1)}, 2021.

\bibitem{berger2013benchmark}
Matthew Berger, Joshua~A Levine, Luis~Gustavo Nonato, Gabriel Taubin, and
  Claudio~T Silva.
\newblock A benchmark for surface reconstruction.
\newblock {\em ACM Transactions on Graphics (TOG)}, 32(2):1--17, 2013.

\bibitem{chang2017matterport3d}
Angel Chang, Angela Dai, Thomas Funkhouser, Maciej Halber, Matthias Niessner,
  Manolis Savva, Shuran Song, Andy Zeng, and Yinda Zhang.
\newblock Matterport3d: Learning from rgb-d data in indoor environments.
\newblock {\em International Conference on 3D Vision (3DV)}, 2017.

\bibitem{shapenet2015}
Angel~X. Chang, Thomas Funkhouser, Leonidas Guibas, Pat Hanrahan, Qixing Huang,
  Zimo Li, Silvio Savarese, Manolis Savva, Shuran Song, Hao Su, Jianxiong Xiao,
  Li Yi, and Fisher Yu.
\newblock {ShapeNet: An Information-Rich 3D Model Repository}.
\newblock Technical Report arXiv:1512.03012 [cs.GR], Stanford University ---
  Princeton University --- Toyota Technological Institute at Chicago, 2015.

\bibitem{VolumeFusion:Choe_2021_ICCV}
Jaesung Choe, Sunghoon Im, Francois Rameau, Minjun Kang, and In~So Kweon.
\newblock Volumefusion: Deep depth fusion for 3d scene reconstruction.
\newblock In {\em Proceedings of the IEEE/CVF International Conference on
  Computer Vision (ICCV)}, pages 16086--16095, October 2021.

\bibitem{corsini2009image}
Massimiliano Corsini, Matteo Dellepiane, Federico Ponchio, and Roberto
  Scopigno.
\newblock Image-to-geometry registration: a mutual information method
  exploiting illumination-related geometric properties.
\newblock In {\em Computer Graphics Forum}, volume~28, pages 1755--1764. Wiley
  Online Library, 2009.

\bibitem{TUM:Cremers-Kolev-pami11}
D. Cremers and K. Kolev.
\newblock Multiview stereo and silhouette consistency via convex functionals
  over convex domains.
\newblock {\em IEEE Transactions on Pattern Analysis and Machine Intelligence},
  33(6):1161--1174, 2011.

\bibitem{curless_volumetric_1996}
Brian Curless and Marc Levoy.
\newblock A volumetric method for building complex models from range images.
\newblock In {\em Proceedings of the 23rd annual conference on {Computer}
  graphics and interactive techniques - {SIGGRAPH} '96}, pages 303--312, Not
  Known, 1996. ACM Press.

\bibitem{ScanNet:dai2017scannet}
Angela Dai, Angel~X Chang, Manolis Savva, Maciej Halber, Thomas Funkhouser, and
  Matthias Nie{\ss}ner.
\newblock Scan{N}et: Richly-annotated {3D} reconstructions of indoor scenes.
\newblock In {\em Proceedings of the IEEE conference on computer vision and
  pattern recognition}, pages 5828--5839, 2017.

\bibitem{SG-NN:Dai_2020_CVPR}
Angela Dai, Christian Diller, and Matthias Niessner.
\newblock Sg-nn: Sparse generative neural networks for self-supervised scene
  completion of rgb-d scans.
\newblock In {\em Proceedings of the IEEE/CVF Conference on Computer Vision and
  Pattern Recognition (CVPR)}, June 2020.

\bibitem{dai_bundlefusion:2017}
Angela Dai, Matthias Nießner, Michael Zollhöfer, Shahram Izadi, and Christian
  Theobalt.
\newblock {BundleFusion}: {Real}-{Time} {Globally} {Consistent} {3D}
  {Reconstruction} {Using} {On}-the-{Fly} {Surface} {Reintegration}.
\newblock {\em ACM Transactions on Graphics}, 36(3):1--18, July 2017.

\bibitem{dai2018scancomplete}
Angela Dai, Daniel Ritchie, Martin Bokeloh, Scott Reed, J{\"u}rgen Sturm, and
  Matthias Nie{\ss}ner.
\newblock Scancomplete: Large-scale scene completion and semantic segmentation
  for 3d scans.
\newblock In {\em Proceedings of the IEEE Conference on Computer Vision and
  Pattern Recognition}, pages 4578--4587, 2018.

\bibitem{SPSG:Dai_2021_CVPR}
Angela Dai, Yawar Siddiqui, Justus Thies, Julien Valentin, and Matthias
  Niessner.
\newblock Spsg: Self-supervised photometric scene generation from rgb-d scans.
\newblock In {\em Proceedings of the IEEE/CVF Conference on Computer Vision and
  Pattern Recognition (CVPR)}, pages 1747--1756, June 2021.

\bibitem{dellepiane2013global}
Matteo Dellepiane and Roberto Scopigno.
\newblock Global refinement of image-to-geometry registration for color
  projection.
\newblock In {\em 2013 Digital Heritage International Congress
  (DigitalHeritage)}, volume~1, pages 39--46. IEEE, 2013.

\bibitem{DeFuSR:Donne_2019_CVPR}
Simon Donne and Andreas Geiger.
\newblock Learning non-volumetric depth fusion using successive reprojections.
\newblock In {\em Proceedings of the IEEE/CVF Conference on Computer Vision and
  Pattern Recognition (CVPR)}, June 2019.

\bibitem{elias2020assessing}
Melanie Elias, Anette Eltner, Frank Liebold, and Hans-Gerd Maas.
\newblock Assessing the influence of temperature changes on the geometric
  stability of smartphone-and raspberry pi cameras.
\newblock {\em Sensors}, 20(3):643, 2020.

\bibitem{germain2020s2dnet}
Hugo Germain, Guillaume Bourmaud, and Vincent Lepetit.
\newblock S2dnet: Learning accurate correspondences for sparse-to-dense feature
  matching.
\newblock {\em arXiv preprint arXiv:2004.01673}, 2020.

\bibitem{gu2020casmvsnet}
Xiaodong Gu, Zhiwen Fan, Siyu Zhu, Zuozhuo Dai, Feitong Tan, and Ping Tan.
\newblock Cascade cost volume for high-resolution multi-view stereo and stereo
  matching.
\newblock In {\em Proceedings of the IEEE/CVF Conference on Computer Vision and
  Pattern Recognition (CVPR)}, June 2020.

\bibitem{ICL-NUIM:Handa_2014}
A. Handa, T. Whelan, J.B. McDonald, and A.J. Davison.
\newblock A benchmark for {RGB-D} visual odometry, {3D} reconstruction and
  {SLAM}.
\newblock In {\em IEEE Intl. Conf. on Robotics and Automation, ICRA}, Hong
  Kong, China, May 2014.

\bibitem{RGB-D-D:He_2021_CVPR}
Lingzhi He, Hongguang Zhu, Feng Li, Huihui Bai, Runmin Cong, Chunjie Zhang,
  Chunyu Lin, Meiqin Liu, and Yao Zhao.
\newblock Towards fast and accurate real-world depth super-resolution:
  Benchmark dataset and baseline.
\newblock In {\em Proceedings of the IEEE/CVF Conference on Computer Vision and
  Pattern Recognition (CVPR)}, pages 9229--9238, June 2021.

\bibitem{herrera2012joint}
Daniel Herrera, Juho Kannala, and Janne Heikkil{\"a}.
\newblock Joint depth and color camera calibration with distortion correction.
\newblock {\em IEEE Transactions on Pattern Analysis and Machine Intelligence},
  34(10):2058--2064, 2012.

\bibitem{DI-Fusion:Huang_2021_CVPR}
Jiahui Huang, Shi-Sheng Huang, Haoxuan Song, and Shi-Min Hu.
\newblock Di-fusion: Online implicit 3d reconstruction with deep priors.
\newblock In {\em Proceedings of the IEEE/CVF Conference on Computer Vision and
  Pattern Recognition (CVPR)}, pages 8932--8941, June 2021.

\bibitem{MSG-Net:Hui2016}
Tak-Wai Hui, Chen~Change Loy, , and Xiaoou Tang.
\newblock Depth map super-resolution by deep multi-scale guidance.
\newblock In {\em Proceedings of European Conference on Computer Vision
  (ECCV)}, pages 353--369, 2016.

\bibitem{izadi_kinectfusion:2011}
Shahram Izadi, Andrew Davison, Andrew Fitzgibbon, David Kim, Otmar Hilliges,
  David Molyneaux, Richard Newcombe, Pushmeet Kohli, Jamie Shotton, Steve
  Hodges, and Dustin Freeman.
\newblock {KinectFusion}: real-time {3D} reconstruction and interaction using a
  moving depth camera.
\newblock In {\em Proceedings of the 24th annual {ACM} symposium on {User}
  interface software and technology - {UIST} '11}, page 559, Santa Barbara,
  California, USA, 2011. ACM Press.

\bibitem{DTU:jensen2014large}
Rasmus Jensen, Anders Dahl, George Vogiatzis, Engil Tola, and Henrik Aan{\ae}s.
\newblock Large scale multi-view stereopsis evaluation.
\newblock In {\em 2014 IEEE Conference on Computer Vision and Pattern
  Recognition}, pages 406--413. IEEE, 2014.

\bibitem{RecScanNet:Jeon2018}
Junho Jeon and Seungyong Lee.
\newblock Reconstruction-based pairwise depth dataset for depth image
  enhancement using cnn.
\newblock In {\em Proceedings of the European Conference on Computer Vision
  (ECCV)}, September 2018.

\bibitem{kazhdan2013screened}
Michael Kazhdan and Hugues Hoppe.
\newblock Screened poisson surface reconstruction.
\newblock {\em ACM Transactions on Graphics (ToG)}, 32(3):1--13, 2013.

\bibitem{kingma2015adam}
Diederik~P. Kingma and Jimmy Ba.
\newblock Adam: {A} method for stochastic optimization.
\newblock In Yoshua Bengio and Yann LeCun, editors, {\em 3rd International
  Conference on Learning Representations, {ICLR} 2015, San Diego, CA, USA, May
  7-9, 2015, Conference Track Proceedings}, 2015.

\bibitem{TNT:knapitsch2017tanks}
Arno Knapitsch, Jaesik Park, Qian-Yi Zhou, and Vladlen Koltun.
\newblock Tanks and temples: Benchmarking large-scale scene reconstruction.
\newblock {\em ACM Transactions on Graphics (ToG)}, 36(4):1--13, 2017.

\bibitem{koch2021hardware}
Sebastian Koch, Yurii Piadyk, Markus Worchel, Marc Alexa, Cl{\'a}udio Silva,
  Denis Zorin, and Daniele Panozzo.
\newblock Hardware design and accurate simulation for benchmarking of {3D}
  reconstruction algorithms.
\newblock In {\em Thirty-fifth Conference on Neural Information Processing
  Systems Datasets and Benchmarks Track (Round 2)}, 2021.

\bibitem{Kuhn2020DeepCMVS}
Andreas Kuhn, Christian Sormann, Mattia Rossi, Oliver Erdler, and Friedrich
  Fraundorfer.
\newblock Deepc-mvs: Deep confidence prediction for multi-view stereo
  reconstruction.
\newblock In {\em 2020 International Conference on 3D Vision (3DV)}, pages
  404--413, 2020.

\bibitem{Leroy_2018_ECCV}
Vincent Leroy, Jean-Sebastien Franco, and Edmond Boyer.
\newblock Shape reconstruction using volume sweeping and learned
  photoconsistency.
\newblock In {\em Proceedings of the European Conference on Computer Vision
  (ECCV)}, September 2018.

\bibitem{ley2016syb3r}
Andreas Ley, Ronny H{\"a}nsch, and Olaf Hellwich.
\newblock Syb3r: A realistic synthetic benchmark for 3d reconstruction from
  images.
\newblock In {\em European Conference on Computer Vision}, pages 236--251.
  Springer, 2016.

\bibitem{lin2021real}
Shanchuan Lin, Andrey Ryabtsev, Soumyadip Sengupta, Brian~L Curless, Steven~M
  Seitz, and Ira Kemelmacher-Shlizerman.
\newblock Real-time high-resolution background matting.
\newblock In {\em Proceedings of the IEEE/CVF Conference on Computer Vision and
  Pattern Recognition}, pages 8762--8771, 2021.

\bibitem{lindenberger2021pixel}
Philipp Lindenberger, Paul-Edouard Sarlin, Viktor Larsson, and Marc Pollefeys.
\newblock Pixel-perfect structure-from-motion with featuremetric refinement.
\newblock In {\em Proceedings of the IEEE/CVF International Conference on
  Computer Vision}, pages 5987--5997, 2021.

\bibitem{liu2021simultaneous}
Yuanzhi Liu, Yujia Fu, Fengdong Chen, Bart Goossens, Wei Tao, and Hui Zhao.
\newblock Simultaneous localization and mapping related datasets: A
  comprehensive survey.
\newblock {\em arXiv preprint arXiv:2102.04036}, 2021.

\bibitem{luo2020attention}
Keyang Luo, Tao Guan, Lili Ju, Yuesong Wang, Zhuo Chen, and Yawei Luo.
\newblock Attention-aware multi-view stereo.
\newblock In {\em Proceedings of the IEEE/CVF Conference on Computer Vision and
  Pattern Recognition}, pages 1590--1599, 2020.

\bibitem{ma2021epp-mvsnet}
Xinjun Ma, Yue Gong, Qirui Wang, Jingwei Huang, Lei Chen, and Fan Yu.
\newblock Epp-mvsnet: Epipolar-assembling based depth prediction for multi-view
  stereo.
\newblock In {\em Proceedings of the IEEE/CVF International Conference on
  Computer Vision (ICCV)}, pages 5732--5740, October 2021.

\bibitem{DeepSurfels:Mihajlovic_2021_CVPR}
Marko Mihajlovic, Silvan Weder, Marc Pollefeys, and Martin~R. Oswald.
\newblock Deepsurfels: Learning online appearance fusion.
\newblock In {\em Proceedings of the IEEE/CVF Conference on Computer Vision and
  Pattern Recognition (CVPR)}, pages 14524--14535, June 2021.

\bibitem{mildenhall2020nerf}
Ben Mildenhall, Pratul~P Srinivasan, Matthew Tancik, Jonathan~T Barron, Ravi
  Ramamoorthi, and Ren Ng.
\newblock Nerf: Representing scenes as neural radiance fields for view
  synthesis.
\newblock In {\em European Conference on Computer Vision}, pages 405--421.
  Springer, 2020.

\bibitem{murez2020atlas}
Zak Murez, Tarrence van As, James Bartolozzi, Ayan Sinha, Vijay Badrinarayanan,
  and Andrew Rabinovich.
\newblock Atlas: End-to-end 3d scene reconstruction from posed images.
\newblock In {\em European Conference on Computer Vision (ECCV)}, 2020.

\bibitem{NYUv2:Silberman2012}
Pushmeet~Kohli Nathan~Silberman, Derek~Hoiem and Rob Fergus.
\newblock Indoor segmentation and support inference from rgbd images.
\newblock In {\em ECCV}, 2012.

\bibitem{DVR:niemeyer2020differentiable}
Michael Niemeyer, Lars Mescheder, Michael Oechsle, and Andreas Geiger.
\newblock Differentiable volumetric rendering: Learning implicit 3d
  representations without 3d supervision.
\newblock In {\em Proceedings of the IEEE/CVF Conference on Computer Vision and
  Pattern Recognition}, pages 3504--3515, 2020.

\bibitem{voxelhashing_url}
Matthias Nie{\ss}ner.
\newblock Voxelhashing.
\newblock \url{https://github.com/niessner/VoxelHashing}.

\bibitem{niessner2013real}
Matthias Nie{\ss}ner, Michael Zollh{\"o}fer, Shahram Izadi, and Marc
  Stamminger.
\newblock Real-time 3d reconstruction at scale using voxel hashing.
\newblock {\em ACM Transactions on Graphics (ToG)}, 32(6):1--11, 2013.

\bibitem{oechsle2021unisurf}
Michael Oechsle, Songyou Peng, and Andreas Geiger.
\newblock Unisurf: Unifying neural implicit surfaces and radiance fields for
  multi-view reconstruction.
\newblock {\em arXiv preprint arXiv:2104.10078}, 2021.

\bibitem{Redwood:Park2017}
Jaesik Park, Qian-Yi Zhou, and Vladlen Koltun.
\newblock Colored point cloud registration revisited.
\newblock In {\em ICCV}, 2017.

\bibitem{pedregosa2011scikit}
Fabian Pedregosa, Ga{\"e}l Varoquaux, Alexandre Gramfort, Vincent Michel,
  Bertrand Thirion, Olivier Grisel, Mathieu Blondel, Peter Prettenhofer, Ron
  Weiss, Vincent Dubourg, et~al.
\newblock Scikit-learn: Machine learning in python.
\newblock {\em the Journal of machine Learning research}, 12:2825--2830, 2011.

\bibitem{unimvsnet_url}
Rui Peng.
\newblock Unimvsnet.
\newblock \url{https://github.com/prstrive/UniMVSNet}.

\bibitem{peng2022rethinking}
Rui Peng, Rongjie Wang, Zhenyu Wang, Yawen Lai, and Ronggang Wang.
\newblock Rethinking depth estimation for multi-view stereo: A unified
  representation.
\newblock In {\em Proceedings of the IEEE/CVF Conference on Computer Vision and
  Pattern Recognition}, pages 8645--8654, 2022.

\bibitem{powell2006newuoa}
Michael~JD Powell.
\newblock The newuoa software for unconstrained optimization without
  derivatives.
\newblock In {\em Large-scale nonlinear optimization}, pages 255--297.
  Springer, 2006.

\bibitem{rakhimov2022npbgpp}
Ruslan Rakhimov, Andrei-Timotei Ardelean, Victor Lempitsky, and Evgeny Burnaev.
\newblock Npbg++: Accelerating neural point-based graphics.
\newblock In {\em Proceedings of the IEEE/CVF Conference on Computer Vision and
  Pattern Recognition}, pages 15969--15979, 2022.

\bibitem{neuralrgbd_url}
Neural RGB-D~Surface Reconstruction.
\newblock Neural rgb-d surface reconstruction.
\newblock
  \url{https://github.com/dazinovic/neural-rgbd-surface-reconstruction}.

\bibitem{riegler2021stable}
Gernot Riegler and Vladlen Koltun.
\newblock Stable view synthesis.
\newblock In {\em Proceedings of the IEEE/CVF Conference on Computer Vision and
  Pattern Recognition}, pages 12216--12225, 2021.

\bibitem{Middlebury14:Scharstein2014}
Daniel Scharstein, Heiko Hirschm{\"u}ller, York Kitajima, Greg Krathwohl, Nera
  Ne{\v{s}}i{\'c}, Xi Wang, and Porter Westling.
\newblock High-resolution stereo datasets with subpixel-accurate ground truth.
\newblock In {\em German Conference on Pattern Recognition}, pages 31--42.
  Springer, 2014.

\bibitem{scharstein2014high}
Daniel Scharstein, Heiko Hirschm{\"u}ller, York Kitajima, Greg Krathwohl, Nera
  Ne{\v{s}}i{\'c}, Xi Wang, and Porter Westling.
\newblock High-resolution stereo datasets with subpixel-accurate ground truth.
\newblock In {\em German conference on pattern recognition}, pages 31--42.
  Springer, 2014.

\bibitem{colmap_url}
Johannes~Lutz Sch\"{o}nberger.
\newblock Colmap.
\newblock \url{https://github.com/colmap/colmap}.

\bibitem{schoenberger2016sfm-colmap}
Johannes~Lutz Sch\"{o}nberger and Jan-Michael Frahm.
\newblock {Structure-from-Motion Revisited}.
\newblock In {\em Conference on Computer Vision and Pattern Recognition
  (CVPR)}, 2016.

\bibitem{schoenberger2016mvs-colmap}
Johannes~Lutz Sch\"{o}nberger, Enliang Zheng, Marc Pollefeys, and Jan-Michael
  Frahm.
\newblock {Pixelwise View Selection for Unstructured Multi-View Stereo}.
\newblock In {\em European Conference on Computer Vision (ECCV)}, 2016.

\bibitem{surfelmeshing_url}
Thomas Schops.
\newblock Surfelmeshing.
\newblock \url{https://github.com/puzzlepaint/surfelmeshing}.

\bibitem{schops2020having}
Thomas Schops, Viktor Larsson, Marc Pollefeys, and Torsten Sattler.
\newblock Why having 10,000 parameters in your camera model is better than
  twelve.
\newblock In {\em Proceedings of the IEEE/CVF Conference on Computer Vision and
  Pattern Recognition}, pages 2535--2544, 2020.

\bibitem{schops_surfelmeshing:2020}
Thomas Schops, Torsten Sattler, and Marc Pollefeys.
\newblock {SurfelMeshing}: {Online} {Surfel}-{Based} {Mesh} {Reconstruction}.
\newblock {\em IEEE Transactions on Pattern Analysis and Machine Intelligence},
  42(10):2494--2507, Oct. 2020.

\bibitem{ETH3D:schops2017multi}
Thomas Schops, Johannes~L Schonberger, Silvano Galliani, Torsten Sattler,
  Konrad Schindler, Marc Pollefeys, and Andreas Geiger.
\newblock A multi-view stereo benchmark with high-resolution images and
  multi-camera videos.
\newblock In {\em Proceedings of the IEEE Conference on Computer Vision and
  Pattern Recognition}, pages 3260--3269, 2017.

\bibitem{MiddleburyMVS:seitz2006comparison}
Steven~M Seitz, Brian Curless, James Diebel, Daniel Scharstein, and Richard
  Szeliski.
\newblock A comparison and evaluation of multi-view stereo reconstruction
  algorithms.
\newblock In {\em 2006 IEEE computer society conference on computer vision and
  pattern recognition (CVPR'06)}, volume~1, pages 519--528. IEEE, 2006.

\bibitem{shim2018gradient}
Inwook Shim, Tae-Hyun Oh, Joon-Young Lee, Jinwook Choi, Dong-Geol Choi, and
  In~So Kweon.
\newblock Gradient-based camera exposure control for outdoor mobile platforms.
\newblock {\em IEEE Transactions on Circuits and Systems for Video Technology},
  29(6):1569--1583, 2018.

\bibitem{shin2019camera}
Ukcheol Shin, Jinsun Park, Gyumin Shim, Francois Rameau, and In~So Kweon.
\newblock Camera exposure control for robust robot vision with noise-aware
  image quality assessment.
\newblock In {\em 2019 IEEE/RSJ International Conference on Intelligent Robots
  and Systems (IROS)}, pages 1165--1172. IEEE, 2019.

\bibitem{shrestha2022real}
Rakesh Shrestha, Siqi Hu, Minghao Gou, Ziyuan Liu, and Ping Tan.
\newblock A real world dataset for multi-view 3d reconstruction.
\newblock In {\em Computer Vision--ECCV 2022: 17th European Conference, Tel
  Aviv, Israel, October 23--27, 2022, Proceedings, Part VIII}, pages 56--73.
  Springer, 2022.

\bibitem{BigBIRD:singh2014bigbird}
Arjun Singh, James Sha, Karthik~S Narayan, Tudor Achim, and Pieter Abbeel.
\newblock {BigBIRD}: A large-scale {3D} database of object instances.
\newblock In {\em 2014 IEEE international conference on robotics and automation
  (ICRA)}, pages 509--516. IEEE, 2014.

\bibitem{SUN_RGBD:song2015sun}
Shuran Song, Samuel~P Lichtenberg, and Jianxiong Xiao.
\newblock {SUN} {RGB-D}: A {RGB-D} scene understanding benchmark suite.
\newblock In {\em Proceedings of the IEEE conference on computer vision and
  pattern recognition}, pages 567--576, 2015.

\bibitem{CA-IRL_SR:Song_2020}
Xibin Song, Yuchao Dai, Dingfu Zhou, Liu Liu, Wei Li, Hongdong Li, and Ruigang
  Yang.
\newblock Channel attention based iterative residual learning for depth map
  super-resolution.
\newblock In {\em Proceedings of the IEEE/CVF Conference on Computer Vision and
  Pattern Recognition (CVPR)}, June 2020.

\bibitem{strecha2008benchmarking}
Christoph Strecha, Wolfgang Von~Hansen, Luc Van~Gool, Pascal Fua, and Ulrich
  Thoennessen.
\newblock On benchmarking camera calibration and multi-view stereo for high
  resolution imagery.
\newblock In {\em 2008 IEEE conference on computer vision and pattern
  recognition}, pages 1--8. Ieee, 2008.

\bibitem{NeuralRecon:Sun2021CVPR}
Jiaming Sun, Yiming Xie, Linghao Chen, Xiaowei Zhou, and Hujun Bao.
\newblock Neuralrecon: Real-time coherent 3d reconstruction from monocular
  video.
\newblock In {\em Proceedings of the IEEE/CVF Conference on Computer Vision and
  Pattern Recognition (CVPR)}, pages 15598--15607, June 2021.

\bibitem{teichman2013unsupervised}
Alex Teichman, Stephen Miller, and Sebastian Thrun.
\newblock Unsupervised intrinsic calibration of depth sensors via slam.
\newblock In {\em Robotics: Science and systems}, volume 248, page~3, 2013.

\bibitem{tibshirani1996regression}
Robert Tibshirani.
\newblock Regression shrinkage and selection via the lasso.
\newblock {\em Journal of the Royal Statistical Society: Series B
  (Methodological)}, 58(1):267--288, 1996.

\bibitem{voynov2019perceptual}
Oleg Voynov, Alexey Artemov, Vage Egiazarian, Alexander Notchenko, Gleb
  Bobrovskikh, Evgeny Burnaev, and Denis Zorin.
\newblock Perceptual deep depth super-resolution.
\newblock In {\em Proceedings of the IEEE/CVF International Conference on
  Computer Vision}, pages 5653--5663, 2019.

\bibitem{wang2021patchmatchnet}
Fangjinhua Wang, Silvano Galliani, Christoph Vogel, Pablo Speciale, and Marc
  Pollefeys.
\newblock Patchmatchnet: Learned multi-view patchmatch stereo.
\newblock In {\em Proceedings of the IEEE/CVF Conference on Computer Vision and
  Pattern Recognition (CVPR)}, 2021.

\bibitem{neus_url}
Peng Wang.
\newblock Neus.
\newblock \url{https://github.com/Totoro97/NeuS}.

\bibitem{wang2021neus}
Peng Wang, Lingjie Liu, Yuan Liu, Christian Theobalt, Taku Komura, and Wenping
  Wang.
\newblock Neus: Learning neural implicit surfaces by volume rendering for
  multi-view reconstruction.
\newblock {\em NeurIPS}, 2021.

\bibitem{wang2021ibrnet}
Qianqian Wang, Zhicheng Wang, Kyle Genova, Pratul~P Srinivasan, Howard Zhou,
  Jonathan~T Barron, Ricardo Martin-Brualla, Noah Snavely, and Thomas
  Funkhouser.
\newblock Ibrnet: Learning multi-view image-based rendering.
\newblock In {\em Proceedings of the IEEE/CVF Conference on Computer Vision and
  Pattern Recognition}, pages 4690--4699, 2021.

\bibitem{CoRBS:wasenmueller2016corbs}
Oliver Wasenm\"uller, Marcel Meyer, and Didier Stricker.
\newblock {CoRBS}: Comprehensive rgb-d benchmark for slam using kinect v2.
\newblock In {\em IEEE Winter Conference on Applications of Computer Vision
  (WACV)}. IEEE, March 2016.

\bibitem{routedfusion_url}
Silvan Weder.
\newblock Routedfusion.
\newblock \url{https://github.com/weders/RoutedFusion}.

\bibitem{weder_routedfusion:2020}
Silvan Weder, Johannes Schonberger, Marc Pollefeys, and Martin~R. Oswald.
\newblock {RoutedFusion}: {Learning} {Real}-{Time} {Depth} {Map} {Fusion}.
\newblock In {\em 2020 {IEEE}/{CVF} {Conference} on {Computer} {Vision} and
  {Pattern} {Recognition} ({CVPR})}, pages 4886--4896, Seattle, WA, USA, June
  2020. IEEE.

\bibitem{NeuralFusion:Weder_2021_CVPR}
Silvan Weder, Johannes~L. Schonberger, Marc Pollefeys, and Martin~R. Oswald.
\newblock Neuralfusion: Online depth fusion in latent space.
\newblock In {\em Proceedings of the IEEE/CVF Conference on Computer Vision and
  Pattern Recognition (CVPR)}, pages 3162--3172, June 2021.

\bibitem{wei2021aa-rmvsnet}
Zizhuang Wei, Qingtian Zhu, Chen Min, Yisong Chen, and Guoping Wang.
\newblock Aa-rmvsnet: Adaptive aggregation recurrent multi-view stereo network.
\newblock In {\em Proceedings of the IEEE/CVF International Conference on
  Computer Vision}, pages 6187--6196, 2021.

\bibitem{whelan_elasticfusion:2016}
Thomas Whelan, Renato~F Salas-Moreno, Ben Glocker, Andrew~J Davison, and Stefan
  Leutenegger.
\newblock {ElasticFusion}: {Real}-time dense {SLAM} and light source
  estimation.
\newblock {\em The International Journal of Robotics Research},
  35(14):1697--1716, Dec. 2016.

\bibitem{ModelNet:CVPR15_Wu}
Zhirong Wu, Shuran Song, Aditya Khosla, Linguang Zhang, Xiaoou Tang, and
  Jianxiong Xiao.
\newblock 3d shapenets: A deep representation for volumetric shape modeling.
\newblock In {\em IEEE Conference on Computer Vision and Pattern Recognition
  (CVPR)}, Boston, USA, June 2015.

\bibitem{MS-PFL:Chuhua2020}
Chuhua {Xian}, Kun {Qian}, Zitian {Zhang}, and Charlie C.~L. {Wang}.
\newblock {Multi-Scale Progressive Fusion Learning for Depth Map
  Super-Resolution}.
\newblock {\em arXiv e-prints}, page arXiv:2011.11865, Nov. 2020.

\bibitem{acmp_url}
Qingshan Xu.
\newblock Acmp.
\newblock \url{https://github.com/GhiXu/ACMP}.

\bibitem{xu2019ACMM}
Qingshan Xu and Wenbing Tao.
\newblock Multi-scale geometric consistency guided multi-view stereo.
\newblock {\em Computer Vision and Pattern Recognition (CVPR)}, 2019.

\bibitem{xu2020ACMP}
Qingshan Xu and Wenbing Tao.
\newblock Planar prior assisted patchmatch multi-view stereo.
\newblock {\em AAAI Conference on Artificial Intelligence (AAAI)}, 2020.

\bibitem{blendedmvg_url}
Yao Yao.
\newblock Blendedmvg.
\newblock \url{https://github.com/YoYo000/BlendedMVS\#upgrade-to-blendedmvg}.

\bibitem{yao2018mvsnet}
Yao Yao, Zixin Luo, Shiwei Li, Tian Fang, and Long Quan.
\newblock Mvsnet: Depth inference for unstructured multi-view stereo.
\newblock {\em European Conference on Computer Vision (ECCV)}, 2018.

\bibitem{BlendedMVS:yao2020blendedmvs}
Yao Yao, Zixin Luo, Shiwei Li, Jingyang Zhang, Yufan Ren, Lei Zhou, Tian Fang,
  and Long Quan.
\newblock Blendedmvs: A large-scale dataset for generalized multi-view stereo
  networks.
\newblock {\em Computer Vision and Pattern Recognition (CVPR)}, 2020.

\bibitem{IDR:yariv2020multiview}
Lior Yariv, Yoni Kasten, Dror Moran, Meirav Galun, Matan Atzmon, Basri Ronen,
  and Yaron Lipman.
\newblock Multiview neural surface reconstruction by disentangling geometry and
  appearance.
\newblock {\em Advances in Neural Information Processing Systems}, 33, 2020.

\bibitem{zacharov2019zhores}
Igor Zacharov, Rinat Arslanov, Maksim Gunin, Daniil Stefonishin, Andrey Bykov,
  Sergey Pavlov, Oleg Panarin, Anton Maliutin, Sergey Rykovanov, and Maxim
  Fedorov.
\newblock ``zhores''-petaflops supercomputer for data-driven modeling, machine
  learning and artificial intelligence installed in skolkovo institute of
  science and technology.
\newblock {\em Open Engineering}, 9(1):512--520, 2019.

\bibitem{zeisl2016structure}
Bernhard Zeisl and Marc Pollefeys.
\newblock Structure-based auto-calibration of rgb-d sensors.
\newblock In {\em 2016 IEEE International Conference on Robotics and Automation
  (ICRA)}, pages 5076--5083. IEEE, 2016.

\bibitem{vismvsnet_url}
Jingyang Zhang.
\newblock Vismvsnet.
\newblock \url{https://github.com/jzhangbs/Vis-MVSNet}.

\bibitem{zhang2020vismvsnet}
Jingyang Zhang, Yao Yao, Shiwei Li, Zixin Luo, and Tian Fang.
\newblock Visibility-aware multi-view stereo network.
\newblock {\em British Machine Vision Conference (BMVC)}, 2020.

\bibitem{DC:Zhang2018}
Yinda Zhang and Thomas Funkhouser.
\newblock Deep {Depth} {Completion} of a {Single} {RGB}-{D} {Image}.
\newblock In {\em 2018 {IEEE}/{CVF} {Conference} on {Computer} {Vision} and
  {Pattern} {Recognition}}, pages 175--185, Salt Lake City, UT, USA, June 2018.
  IEEE.

\bibitem{zhou2014simultaneous}
Qian-Yi Zhou and Vladlen Koltun.
\newblock Simultaneous localization and calibration: Self-calibration of
  consumer depth cameras.
\newblock In {\em Proceedings of the IEEE Conference on Computer Vision and
  Pattern Recognition}, pages 454--460, 2014.

\bibitem{zhou2018open3d}
Qian-Yi Zhou, Jaesik Park, and Vladlen Koltun.
\newblock {Open3D}: {A} modern library for {3D} data processing.
\newblock {\em arXiv:1801.09847}, 2018.

\end{thebibliography}
    }

%
\end{document}